To appear in *Knowledge and Information Systems*, 2023.

# The impact of prior knowledge on causal structure learning

Anthony C. Constantinou, Zhigao Guo, and Neville K. Kitson.

Bayesian AI research lab, Machine Intelligence and Decision Systems (MInDS) research group, School of Electronic Engineering and Computer Science, Queen Mary University of London (QMUL), London, UK, E1 4NS.

E-mails: a.constantinou@qmul.ac.uk, zhigao.guo@qmul.ac.uk, and n.k.kitson@qmul.ac.uk.

**Abstract:** Causal Bayesian networks have become a powerful technology for reasoning under uncertainty in areas that require transparency and explainability, by relying on causal assumptions that enable us to simulate hypothetical interventions. The graphical structure of such models can be estimated by structure learning algorithms, domain knowledge, or a combination of both. Various knowledge approaches have been proposed in the literature that enable us to specify prior knowledge that constrains or guides these algorithms. This paper introduces some novel, and also describes some existing, knowledge-based approaches that enable us to combine structure learning with knowledge obtained from heterogeneous sources. We investigate the impact of these approaches on structure learning across different algorithms, case studies and settings that we might encounter in practice. Each approach is assessed in terms of effectiveness and efficiency, including graphical accuracy, model fitting, complexity, and runtime; making this the first paper that provides a comparative evaluation of a wide range of knowledge approaches for structure learning. Because the value of knowledge depends on what data are available, we illustrate the results both with limited and big data. While the overall results show that knowledge becomes less important with big data due to higher learning accuracy rendering knowledge less important, some of the knowledge approaches are found to be more important with big data. Amongst the main conclusions is the observation that reduced search space obtained from knowledge does not always imply reduced computational complexity, perhaps because the relationships implied by the data and knowledge are in tension.

## 1. Introduction

Learning Bayesian Networks (BNs) involves constructing a graphical structure and parameterising its conditional distributions given the structure. The structure of a BN, also referred to as Causal BN (CBN) if the arcs are assumed to represent causation, can be determined by causal knowledge, learnt from data, or a combination of both. Problems with access to domain expertise tend to attract manual construction of CBN models, whereas automated construction is more prevalent in problems where knowledge is limited or highly uncertain, and structure learning is expected to provide insights that would otherwise remain unknown.

Automated causal discovery is hindered by difficulties that have significantly limited its impact in practise. Algorithms will often discover incorrect causal relationships for events that humans perceive as common sense. For example, an algorithm may discover that the music is influenced by the moves of the dancer, rather than concluding that the dancer is dancing to the music. Such counterintuitive causal links represent failures of causal common-sense and raise questions as to whether machine learning is capable of achieving human level causal understanding.

There are several important reasons that make causal discovery difficult. These include that it is not possible to learn all causal relationships from observational data alone, since the best an algorithm can do is to recover the causal structure up to its Markov equivalence class when learning from observational data. Moreover, the satisfactory performance the algorithms tend to achieve in clean synthetic experiments is rarely repeated in practice [17], and this is because real data are imperfect and tend to violate many of the assumptions on which these algorithms rely on, such as complete data, causal sufficiency, and error-free data, amongst others. Last but not least, exploring the search space of graphs is, in general, NP-hard.

Discovering reasonably accurate causal structure is very important in today's world, where automated decision-making is increasingly being questioned if it provides no transparency or explainability. Unlike black-box solutions, Pearl's $do$-calculus framework that relies on CBNs [39, 40] provides a clear mechanism for computing the effect of hypothetical interventions. In other words, we can use an optimisation function to determine the optimal intervention by manipulating variables within a causal model in a way that renders the manipulation of a given variable independent of its causes, thereby generating the effect of intervention only. While $do$-calculus is a relatively simple and a powerful tool for simulating experimental data, its effectiveness directly relies on the accuracy of the underlying causal structure of the model; implying that structure learning accuracy has a direct effect on the accuracy of intervention.

If we expect algorithms to become rational modellers of a world that requires causal perception, then we may have to provide them with something more than an imperfect static observational data set. Structure learning constraints that embody causal knowledge represent one way of assisting algorithms to produce more accurate causal graphs. Because it is common to have some understanding of the underlying causal mechanisms of a given domain, practitioners often seek access to expert knowledge that can guide or constrain structure learning, effectively reducing the uncertainty of the BN models learnt from data.

This paper introduces some novel, and also describes some existing, knowledge-based approaches to investigate their effectiveness on structure learning, by applying them to multiple algorithms and case studies over varying degrees of data size and quantity of knowledge constraints. The paper is structured as follows: Section 2 provides a review of past relevant studies, Section 3 describes the knowledge approaches we have implemented, Section 4 describes the algorithms we have modified to enable knowledge-based constraints to be incorporated into their structure learning process, Section 5 describes the simulation process and experiments, we present and discuss the results in Section 6, and concluding remarks are given in Section 7.

## 2. Review of past relevant studies

The field of causal structure learning has seen important advances over the past few decades, particularly those coming from the constraint-based and score-based classes of learning. The relevant literature has evolved rapidly with innovations involving both constraint-based and score-based learning, which represent the two main classes of structure learning. Kitson et al. [30] provide an extensive survey of these algorithms.

The main difference between these two classes of learning is that constraint-based algorithms, such as the popular PC [47] and FCI [48] algorithms and their stable versions [13], tend to start with a fully connected graph and use conditional independence tests to eliminate and orientate some of those edges. On the other hand, score-based algorithms which include the well-established Greedy Equivalence Search (GES) by Chickering and Meek [11], Hill Climbing (HC) and TABU search [3, 27, 44], typically start from an empty graph and iteratively modify the graph to increase its objective score. In this paper, our focus will be on score-based algorithms, but also hybrid learning algorithms that combine the two above classes of learning.

There are different types of score-based algorithms. An important distinction between them involves whether the algorithm offers approximate or exact learning solutions. Exact score-based algorithms such as GOBNILP [20] and B&B [22] tend to be based on combinatorial optimisation approaches in conjunction with sound pruning strategies that reduce the scale of combinatorial optimisation. In contrast, most approximate algorithms, including GES, HC and TABU discussed above, tend to rely on heuristics that iteratively traverse the search space of graphs through arc additions, reversals and removals, and move towards the graph that maximises a given objective function. While exact solutions guarantee to return the graph that maximises an objective function, this guarantee is subject to certain data assumptions and subject to limiting the maximum in-degree. Moreover, the use of exact learning approaches is typically limited to smaller networks due to their relatively high computational complexity. On the other hand, approximate algorithms are generally much faster and aim to return a high scoring graph that tends to get stuck at a local optimum, without limiting the maximum in-degree. In this paper, we will be focusing on such approximate learning solutions.

The need to account for causal knowledge when constructing or learning causal representation models is well documented [23]. There is also a lot of relevant work in the field of cognitive psychology on how causal induction constrains or guides human learning. For example, Tenenbaum et al. [49] propose a framework which they call *causal grammar*, about how we think in terms of causal representation and inference, and outlined a formal approach towards solving causal induction inspired by hierarchical Bayesian models. Griffiths and Tenenbaum [25] further build on this view and present a computational framework for the explanation of human causal induction. They use this to identify key aspects of abstract prior knowledge, to show how they provide the constraints that people may rely on when inferring causal structure, and apply it to different case studies to illustrate how explanations of causal learning can be presented. There is also work on how people choose interventions to learn causal systems. This includes the papers by Coenen et al. [12] who show that people make decisions in an adaptive fashion where a trade-off is made between expected performance and cognitive effort, and by Bramley et al. [4] who propose a scheme to understand how people choose interventions to identify causation.

In the field of causal machine learning, many structure learning algorithms can be modified to account for causal knowledge that guides search towards a preferred causal structure, or restricts the search space of graphs that can be explored to those in which graphs contain or do not contain certain relationships. A knowledge approach that guides, rather than restricts, structure learning is often referred to as an approach that introduces soft constraints, whereas an approach that restricts learning towards specific graphs tends to be referred to as an approach that introduces hard constraints.

The work by Heckerman and Geiger [26] can be viewed as one of the earliest implementations of a soft constraint approach, where the structure learning process is modified to reward graphs that are closer to an initial best guess knowledge graph. In other studies, soft constraints are often implemented by assigning prior beliefs as probabilities to some of the possible relationships between pairs of variables, as described by Castelo and Siebes [7]. Likewise, the work of Amirkhani et al. [1] explored the same types of prior probabilities on edge inclusion or exclusion from multiple experts, applied to two variants of the score-based HC algorithm; one of the early implementations of HC in Bayesian Network Structure Learning (BNSL) can be found in [2]. Ibrahim et al. [28] looked at the problem of learning causal structure by augmenting the distribution of causal structures with a user model whose parameters account for possible biases of the expert. BCano et al. [6] presented a methodology that reduces the effort associated with knowledge elicitation by restricting the elicitation of prior knowledge to edges that cannot be reliably discerned from data, in addition to eliciting knowledge about the temporal order of the variables (this information was required by the algorithm), with application to structure learning via MCMC simulation. Masegosa and Moral [37] extended the work of Cano et al. [6] by removing the requirement of knowledge about the temporal order of the variables, thereby reducing elicitation effort at the expense of the algorithm exploring a larger search space of graphs. For practical reasons, the type of soft constraints that we consider in this paper involve those which do *not* require the user to assign probabilities.

Hard constraints, on the other hand, tend to be more popular and often easier to implement. One of the first implementations of hard constraints involves the score-based K2 algorithm [19] which assumes that the temporal order of the data variables is given, thereby restricting the search space of graphs to those consistent with the temporal ordering. Other relevant studies include the work of de Campos and Castellano [21] on measuring the effect of knowledge about the existence or absence of edges, in addition to node ordering constraints, with application to the score-based HC and constraint-based PC [48] algorithms. Incomplete temporal orderings, which involve providing temporal information for some of the variables available in the data were investigated by Ma et al. [36] with application to score-based learning and model-averaging. Chen et al. [10] investigated constraining edges in terms of ancestral, rather than parental, relationships and with application to score-based algorithms with a decomposable[1] objective function, despite the ancestral constraints being non-decomposable; i.e., the ancestral relationships cannot be assigned independent scores in the same way the parental relationships can. Ancestral constraints were also investigated by Li and van Beek [33] who showed that algorithms using non-decomposable ancestral constraints can scale up to problems of 50 variables, with application to the score-based MINOBS algorithm [32] which is an approximate learning algorithm that otherwise scales up to thousands of variables. The work of Chen et al. [10] was recently further investigated by Wang et al. [50] who extended ancestral constraints to score-based learning that searches in the space of variable orderings. Lastly, Constantinou [15] introduced the assumption that all the variables that make up the input data are relevant, which in turn imposes a constraint on the learnt graph not to contain independent subgraphs or unconnected variables, to ensure the learnt model enables full propagation of evidence when applied in practice. Readers may also find helpful the works by Bramley et al. [5] and Rehder et al. [41], who study the role of time in causal learning in humans. The former study illustrates how participants can use the order in which events occurred to narrow down candidate causal structures, and the latter extends this idea to dynamic causal systems where causes continually influence their effects.

Structure learning freeware such as Tetrad [8, 9], bnlearn [43] and CaMML [31, 38] provide access to a range of soft and hard knowledge constraints that can guide or constrain structure learning. To the best of our knowledge, Tetrad is the structure learning freeware that supports the widest range of approaches that enable knowledge to be combined with data for causal discovery. All three, bnlearn, CaMML and Tetrad enable users to specify knowledge about required as well as forbidden edges. Tetrad also allows users to specify the temporal order of the variables, while CaMML enables users to specify a best-guess input graph [31, 38] as in [26].

While it seems plausible that the best approach to constructing a causal graph involves combining knowledge with machine learning, most real-world BN models published in the literature continue to rely entirely on knowledge or entirely on automation, despite the above efforts. Therefore, it is fair to say that the methods that combine knowledge with structure learning remain underused in practice. We often see applications of BNs that either rely on full automation, or entirely on expert systems, rather than combining these two approaches towards causal representation. Perhaps one of the reasons is because their combined effectiveness has not been adequately researched, which partly motivates this study.

## 3. The knowledge approaches implemented

This section discusses and describes the knowledge approaches (both novel and existing) we have implemented in this study. All the knowledge approaches are made available in the Bayesys open-source BNSL system [14].

We begin by distributing and discussing the knowledge approaches we have implemented into relevant categories, and with reference to Table 1 which provides the complete list and description of these knowledge approaches. The categories are:

[1] A decomposable objective function refers to a scoring process that can be decomposed into independent components. In the context of BNSL, a decomposable objective function assigns a score to each node in the graph, where the overall score of the graph is the sum over all individual node scores.

**Table 1.** The knowledge approaches implemented in Bayesys [14] as part of this research paper.

| ID | Knowledge approach | Knowledge input example | Knowledge | Constrains or guides |
|---|---|---|---|---|
| DIR-EDG | Directed edge | $A \rightarrow B$ | Causal relationship or direct dependency. | Constrains the search space of graphs to those containing $A \rightarrow B$. |
| UND-EDG | Undirected edge | $A - B$ | Causal relationship or direct dependency without knowledge of the direction of the relationship. | Constrains the search space of graphs to those containing $A \rightarrow B$ or $A \leftarrow B$. |
| FOR-EDG | Forbidden edge | $A \perp B$ | No causal relationship or direct dependency. | Constrains the search space of graphs to those not containing $A \rightarrow B$ and $A \leftarrow B$. |
| REL-TEM | Relaxed incomplete temporal order | Tier 1: $\{A\}$<br>Tier 2: $\{B, C\}$ | Temporal information such that $B$ and $C$ occur after observing $A$ and hence, $B$ and $C$ cannot be parents nor ancestors of $A$. | Constrains the search space of graphs to those not containing $A \leftarrow B, A \leftarrow C$, or $B$ and/or $C$ as ancestors of $A$. |
| STR-TEM | Strict incomplete temporal order | Tier 1: $\{A\}$<br>Tier 2: $\{B, C\}$ | In addition to constraint $REL - TEM$, it prohibits edges between nodes of the same tier. | As in REL-TEM, plus constrains the search space of graphs to those not containing $B \rightarrow C$ or $B \leftarrow C$. |
| INI-GRA | Initial graph | DAG | An initial best guess graph | Can be viewed as a soft constraint that guides structure learning by changing the starting point in the search space of graphs from an empty graph to an initial best-guess graph. |
| VAR-REL | Variables are relevant | n/a | All variables in the input data are relevant. | The learnt graph must not contain disjoint subgraphs or unconnected nodes. |
| TAR-VAR | Target variable/s | A node $A$ or a set of nodes $\{A, B\}$ | A target variable, or a set of variables, for which we would favour higher dimensionality in the form of an increased parent-set for those variables. | Can be viewed as a soft constraint that guides the search space of graphs to those in which targeted variable $A$, or variable set $\{A, B\}$, is assigned a relaxed dimensionality penalty, as determined by the user, through the objective function. |
| REL-BDN | Relaxed BDNs | A set of Decision nodes $\{A, B\}$ and Utility nodes $\{C, D\}$ | Some of the input variables represent decisions or utilities. | The learnt graph is a BDN graph containing Decisions $\{A, B\}$ and Utilities $\{C, D\}$ represented by rectangle and diamond nodes respectively, where arcs entering Decisions are Informational represented by a dashed arc. |
| STR-BDN | Strict BDNs | A set of Decision nodes $\{A, B\}$ and Utility nodes $\{C, D\}$ | | As in REL-BDN, plus constrains the learnt graphs to those where Decision nodes $\{A, B\}$ and Utility nodes $\{C, D\}$ have at least one child and parent node respectively. |

a) **Direct relationships:** this category represents existing approaches that capture information about direct relationships between nodes. Approaches *Directed edge*[2] (DIR-EDG) and *Forbidden edge*[3] (FOR-EDG), represent the most commonly used knowledge approach. On the other hand, *Undirected edge* (UND-EDG) is less common and indicates knowledge of an edge without knowledge about the direction of the edge.

b) **Incomplete temporal order**: we based our temporal definitions on those in Tetrad where variables are assigned to temporal tiers that represent a timeline. Variables assigned to higher-numbered tiers cannot be parents, but can be indirect causes such as ancestors, of variables assigned to lower-numbered tiers, with the option to prohibit arcs between variables that belong to the same tier. We specify the approaches *Relaxed incomplete temporal order* (REL-TEM) which prohibits edges between variables that violate the temporal tiers, and *Strict incomplete temporal order* (STR-TEM) which prohibits edges between variables that are within the same tier, in addition to what is prohibited by REL-TEM, as described in Table 1.

   The term 'incomplete' implies that not all variables need to be specified in the temporal ordering, and those not included are not subject to any temporal restrictions. In implementing this approach, we found that limiting temporal constraints to parental relationships can be problematic. This is because graphs such as $A \rightarrow B \rightarrow C$ do not violate, for example, ordering $\{t_1 = C, t_2 = A\}$ when $A$ is specified to occur only after observing $C$. That is, while the temporal ordering does not allow $A$ to serve as a parent of $C$, it allows $A$ to serve as an ancestor[4] of $C$. On this basis, we have extended the definition of temporal constraints to include ancestral relationships, in addition to parental relationships. This extended definition is applied to both REL-TEM and STR-TEM approaches. Fig 1 presents a set of examples illustrating under what circumstances arcs violate the ancestral, but not the parental, temporal ordering.

   Although computationally less efficient, incomplete temporal order provides flexibility with knowledge elicited, which is why we explore this approach rather than an approach that requires that the full temporal order is given. For example, it would be unreasonable to force the user to provide full knowledge about the ordering of the variables, as this might substantially increase the knowledge elicitation effort as well as the risk of eliciting inaccurate knowledge.

[2] This is referred to as *Required edge* in Tetrad.

[3] This is identical to the TETRAD implementation but differs from bnlearn's blacklist function which can be viewed as a function that enables users to specify *Forbidden Directed Edge.*

[4] If a complete node ordering is specified, then graphs consistent with that ordering will automatically be consistent with the ancestral relationships in that ordering. This applies to algorithms such as K2, but also to algorithms such as MINOBS which operate in the space of variable orderings even though an ordering is not given as an input. On the other hand, ancestral constraints need to be checked when an incomplete temporal ordering is given, and checking for ancestral relationships represents a computationally expensive task (Chen et al., 2016; Li and van Beek, 2018; Wang et al., 2021), as we also later illustrate in Section 6. This also explains why order-based algorithms designed to handle thousands of variables operate on complete, and not on incomplete, variable orderings.

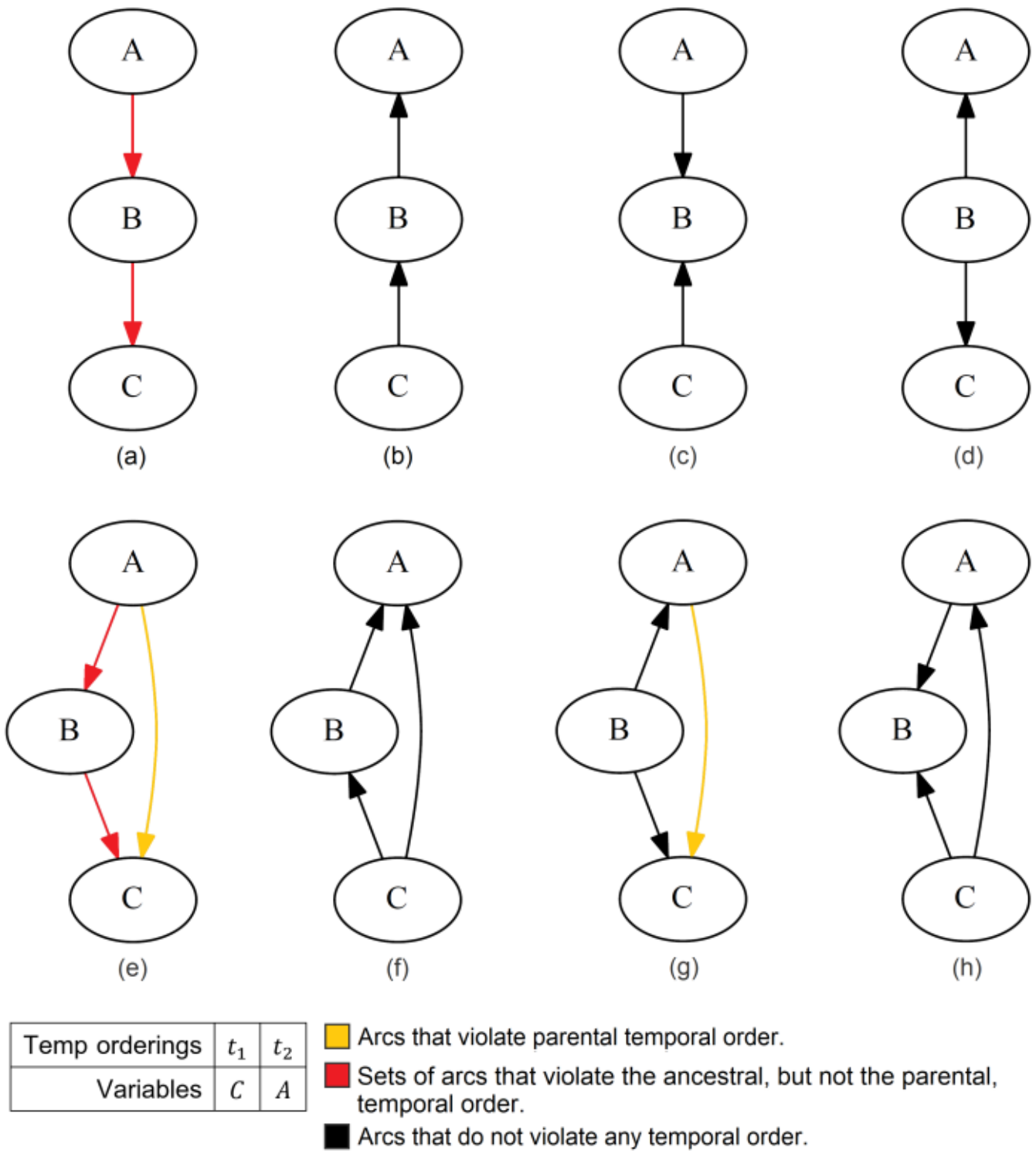

**Fig 1.** A set of different DAGs highlighting the arcs that violate the temporal constraints indicated in the bottom left corner, with reference to the two different types of incomplete temporal order.

c) **Initial graph:** this represents a far less common approach that involves guiding structure learning from a given best-guess initial graph, rather than from an empty or a fully connected graph (refer to INI-GRA in Table 1). The motivation here is to guide a structure learning algorithm into a potentially higher-scoring local maxima. In this paper, we define this approach as an initial graph that serves as the starting point in the search space of DAGs and hence, it involves guiding structure learning *from* the input graph rather than *towards* the input graph.

d) **Variables are relevant:** this can be viewed as knowledge that the variables that make up the input data are relevant and hence, it imposes a constraint on the learnt graph not to contain independent subgraphs or unconnected variables (refer to VAR-REL in Table 1), similar to the structure learning feature in [15] discussed in Section 2. In this paper, this approach introduces additional arcs at the end of a structure learning process, where the additional arcs minimally decrease the given objective function, and this process continues until no independent subgraphs or nodes are present in the learnt graph.

e) **Target node/s:** this represents a novel approach that encourages structure learning to produce a higher number of parents, in this case viewed as potential causes, of targeted variables of interest (refer to TAR-VAR in Table 1). While this approach is not necessarily meant to improve the accuracy of a learnt graph, it can still be useful when working with high dimensional real data of limited sample size. This is because, in those scenarios, structure learning algorithms tend to produce sparse networks that represent an inadequate indication of the potential causes or parent nodes of targeted variables of interest.

This approach encourages structure learning to produce parent-sets of a higher size for targeted nodes by modifying the objective function to scale down the dimensionality penalty associated with the number of free parameters, as to encourage the discovery of more parents for given target nodes. Essentially, this approach exchanges a faster increase in dimensionality for a corresponding smaller increase in Log-Likelihood ($LL$). In this paper, we employ the Bayesian Information Criterion (BIC) as the objective function for graph $G$ and data $D$, modified as $BIC_{TAR-VAR}$ to accept optional adjustments in the dimensionality penalty of a

given target variable:

$$BIC_{TAR-VAR} = LL(G|D) - \left(\frac{log_2 N}{2}\right)p \tag{1}$$

where $N$ is the sample size of $D$, and $p$ is the number of free parameters in $G$, given

$$p = \sum_{i}^{|V|}\left((s_i - 1)\prod_{j}^{|\pi_{v_i}|} q_j\right) \Bigg/ r_i \tag{2}$$

where $V$ is the set of variables in graph $G$, $|V|$ is the size of set $V$, $s_i$ is the number of states of $v_i$, $\pi_{v_i}$ is the parent set of $v_i$, $|\pi_{v_i}|$ is the size of set $\pi_{v_i}$, $q_j$ is the number of states of $v_j$ in parent set $\pi_{v_i}$, and $r_i$ is the new parameter used to diminish the free parameters penalty for targeted variables; i.e., $r = 1$ for non-targeted variables and $r > 1$, as determined by the user, for targeted variables.

f) **Bayesian Decision Networks (BDNs):** this category involves knowledge needed to produce graphs that can be parameterised and used as a BDN, also known as Influence Diagrams (IDs) [46]. While this category is often unrelated to structure learning, we have taken advantage of some of the knowledge needed to convert a BN into a BDN, to devise structural constraints. On this basis, we present this category as an additional knowledge approach that can be used to further constrain structure learning.

BDNs are extensions of BNs suitable for optimal decision-making based on the maximum expected utility criterion, and include additional types of nodes and edges, as well as some inferential constraints. In addition to the uncertain *chance* nodes, BDNs contain *decision* nodes represented by rectangles (these capture the available decision options), and *utility* nodes represented by diamonds (these represent the variable whose value we want to minimise or maximise). BDNs also include *Information* arcs, in addition to conditional arcs, that enter decision nodes and indicate that a decision is rendered independent of its parent nodes. That is, decision options do not represent observations that need to be explained by their causes, although decisions *are* informed by their causes. Dashed arcs are used to indicate that a decision is informed by its parents. This is illustrated in Fig 2 that presents a hypothetical BDN where decision nodes ‘Test’ and ‘Treatment’ are informed by, although rendered independently of, ‘Symptoms’ and ‘TestResult’. In this example, the best decision combination is the one that collectively minimises ‘SideEffect’ and maximises post-treatment effectiveness on ‘DisPostTreat’ and ‘SymPostTreat’.

This means that information entailed by BDNs about *decision* and *utility* nodes can be used to impose some constraints during structure learning. Specifically, a decision node *must* have at least one child node, and a utility node *must* have at least one parent node. This information can be converted into structural constraints, which we later illustrate in Section 6 (refer to Fig 13). The implementation of BDN-based knowledge approaches is separated into two versions, which we define as *Relaxed BDNs* (REL-BDN) and *Strict BDNs* (STR-BDN) in Table 1. Approach REL-BDN represents the visual modifications needed to convert a BN into a BDN for a given set of decision and utility nodes, whereas STR-BDN imposes the graphical constraints discussed above. While BDNs have been around for a while, they have always involved manual construction and, to the best of our knowledge, this is the first study to automate their graphical construction.

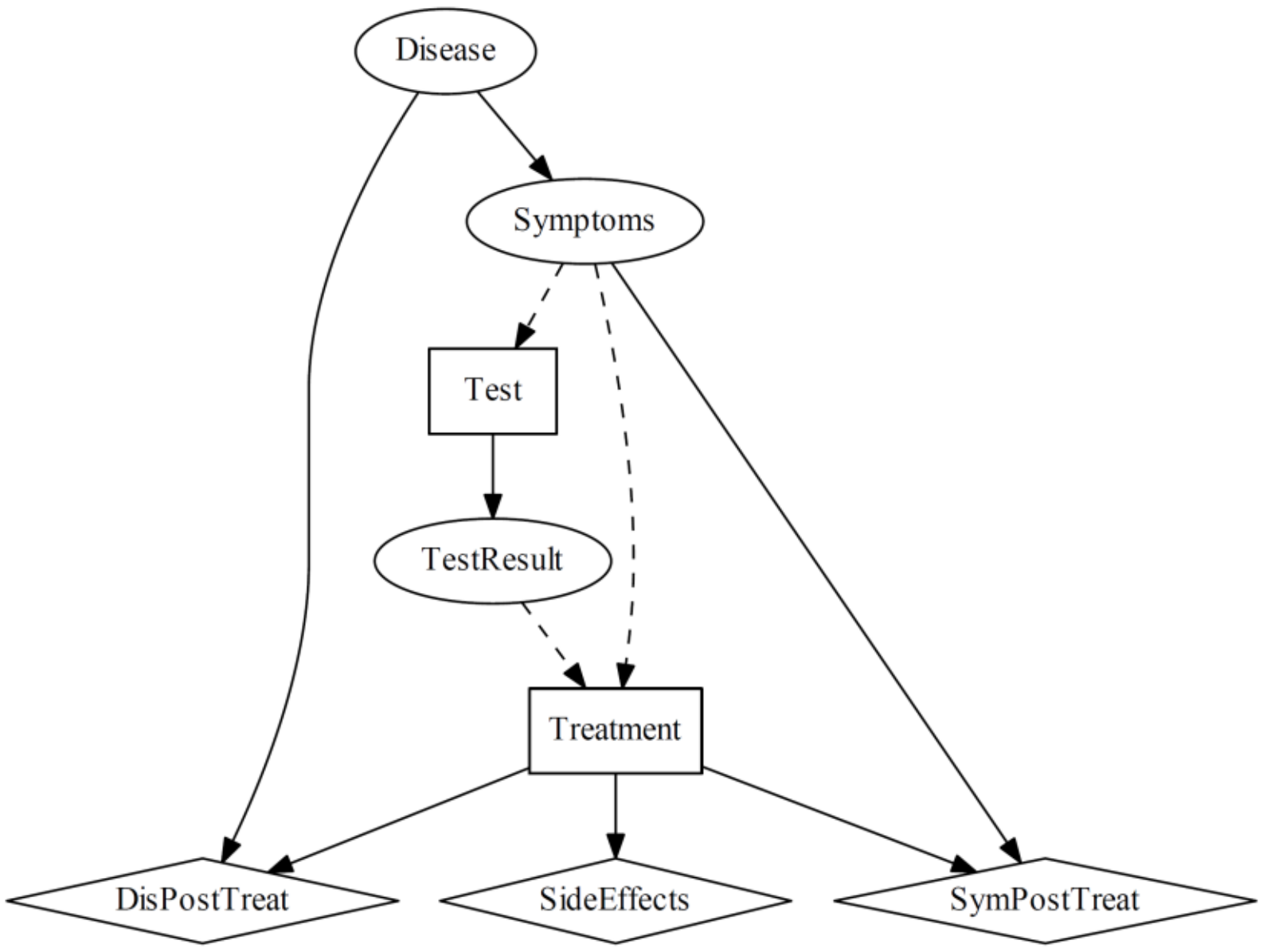

**Fig 2.** An example of an Influence Diagram/Bayesian Decision Network, inspired by Yet et al. [51, 52], where treatment decision is determined by expected utilities on side effects, post-treatment disease, and post-treatment symptoms.

## 4. The algorithms modified to support knowledge-based constraints

This section describes the five algorithms we have modified to enable them to account for the different types of knowledge constraints. The implementation of all five algorithms is made available in the Bayesys open-source BNSL system [14].

Specifically, we modified the five BNSL algorithms to be able to consider information related to each of the 10 knowledge approaches described in Section 3. The five algorithms are:

i. **HC:** The Hill-Climbing algorithm represents the simplest as well as one of the earliest score-based algorithms for structure learning [2, 27]. The starting point of the algorithm is typically an empty or a random graph. In this study, we start from an empty graph to eliminate randomisation, but also to be able to impose the knowledge constraints. HC investigates all possible neighbouring acyclic graphs by performing arc additions, reversals and removals at each iteration. It then applies the graph modification that maximises the objective score, and stops when the objective score no longer increases. The HC is an approximate algorithm that will often get stuck at a local maximum.

ii. **TABU:** This algorithm can be viewed as an extension of HC that permits exploring graphical modifications that decrease the objective score, and maintains a *tabu* list containing the most recently visited graphs to prevent the algorithm from returning to a graph that was previously visited [3]. In practice, this approach enables TABU to move into new graphical regions that may contain an improved local maximum. Because TABU cannot guarantee to return the highest scoring graph available in the search space, it is also an approximate learning algorithm.

iii. **SaiyanH:** This is a hybrid learning algorithm since it relies on both constraint-based and score-based learning strategies [15]. SaiyanH starts by drawing undirected edges between nodes that have the strongest marginal dependency, and then orientates those edges by performing a sequence of conditional independence tests, score-based tests, as well as exploring which orientations would maximise the effect of hypothetical interventions. It then performs *tabu* search with the restriction not to visit graphs that contain disjoint subgraphs. This learning strategy ensures that the learnt graph enables full propagation of evidence when then graph is parameterised.

iv. **MAHC:** The Model-Averaging Hill-Climbing algorithm combines two learning strategies with hill-climbing search [18]. MAHC starts by pruning the search space of graphs. The pruning strategy restricts the candidate parents for each node, and represents an aggressive version of those that are typically applied to combinatorial optimisation structure learning problems. Once a reduced search-space is obtained, it performs model averaging in the hill-climbing search process and moves to the neighbouring graph that maximises the average objective score, across that neighbouring graph and all *its* valid neighbouring graphs. The motivation here is that a model averaging process might be less sensitive to data noise, which is commonly present in real data.

v. **GES:** The Greedy Equivalence Search (GES) algorithm contains two learning phases, known as the forward and backward search phases [11]. The forward phase starts from an empty graph and, at each iteration, the edge that maximises the objective function is added to the graph. When no edge is found to further increase the objective score, GES enters the backward phase where edge removals are performed in the same way. The algorithm stops when no further edge removals increase the objective score.

To be able to measure the impact of knowledge consistently across all algorithms, we employ the same objective function, the BIC, for all algorithms. Moreover, because the knowledge approaches are implemented such that they constrain or guide the search-space of DAGs, we apply all algorithms to the search-space of DAGs. This also applies to GES which is proposed for CPDAG-space exploration; i.e., we had to simplify GES by applying it to the search-space of DAGs without testing for score equivalence. This simplification was necessary since testing for equivalence in the DAG-space (as opposed to the smaller CPDAG-space in [11]) proved to be computationally expensive and led to most experiments failing to complete learning within the runtime limit.

Algorithms 1, 2, 3, 4, and 5 describe how each of these five algorithms was modified to account for the knowledge approaches enumerated in Table 1. Note that temporal knowledge is imposed either as REL-TEM or STR-TEM, and the same applies for knowledge relating to BDNs which is imposed either as REL-BDN or STR-BDN.

Each knowledge approach involves additional constraints that can be taken into consideration by a structure learning algorithm. This means that whenever a knowledge approach appears in the pseudocode of the five algorithms, it implies that the learning is constrained by the information present in the knowledge. As an example, Table 2 presents the constraints for DIR-EDG considered at the 20% rate of constraints for case study Alarm, and this format is followed by all the approaches. For example, the approach UND-EDG would involve columns referring to Var1 and Var2, whereas approach REL-BDN/STR-BDN involves data rows referring to Decision and Utility nodes, rather than Parent and Child nodes. The only approach that involves a rather different structural format is the REL-TEM/STR-TEM, which involves temporal constraints as configured in Table 3.

**Table 2.** An example of directed constraints based on the 20% rate of constraints for case study Alarm. The table shows how this information is encoded into an input data file which is read by an algorithm whenever the knowledge approach DIR-EDG is called.

| Constraint ID | Parent | Child |
|---|---|---|
| 1 | ERRCAUTER | HRSAT |
| 2 | LVFAILURE | LVEDVOLUME |
| 3 | TPR | CATECHOL |
| 4 | LVEDVOLUME | PCWP |
| 5 | LVFAILURE | STROKEVOLUME |
| 6 | VENTMACH | VENTTUBE |
| 7 | INTUBATION | SHUNT |
| 8 | INTUBATION | PRESS |
| 9 | SAO2 | CATECHOL |

**Table 3.** An example of temporal constraints based on the 20% rate of constraints for case study Alarm. The table shows how this information is encoded into an input data file which is read by an algorithm whenever the knowledge approach REL-TEM or STR-TEM is called.

| ID | Tier 1 | Tier 2 | Tier 3 | Tier 4 | Tier 5 | Tier 6 |
|---|---|---|---|---|---|---|
| 1 | VENTMACH | ARTCO2 | EXPCO2 | LVFAILURE | ERRLOWOUTPUT | HRSAT |
| 2 | | | | HYPOVOLEMIA | | |

The implementation of the knowledge approaches can be summarised as follows:

a) **HC, TABU, MAHC and GES:** Their initial empty graph can be replaced by INI-GRA or modified to incorporate DIR-EDG and UND-EDG;
b) **SaiyanH:** Its graph obtained by associational and constraint-based learning can be restricted to graphs that do not violate FOR-EDG, DIR-EDG, UND-EDG, REL-TEM or STR-TEM, and INI-GRA;
c) **MAHC:** Its pruning strategy can be restricted to Candidate Parent Sets (CPSs) that do not violate FOR-EDG, DIR-EDG, and REL-TEM or STR-TEM;
d) **All five algorithms:** Their heuristic search is restricted to graphs that do not violate FOR-EDG, DIR-EDG, UND-EDG, and REL-TEM or STR-TEM;
e) **All five algorithms:** Their objective function is modified to account for TAR-VAR;
f) **HC, TABU, MAHC and GES:** When these four algorithms complete learning, they can be forced to continue the heuristic search until the VAR-REL condition is met;
g) **All five algorithms:** When they complete learning, their graph can be converted into a BDN to satisfy REL-BDN;
h) **All five algorithms:** When they complete learning, they can be forced to extend the heuristic search until the learnt graph satisfies STR-BDN.

**Algorithm 1:** Hill-Climbing (HC) algorithm with additional pseudocode needed to account for the knowledge approaches highlighted in red font colour.

**Input:** dataset $D$, empty graph $G$, objective function $\text{BIC}_{TAR-VAR}(G,D)$, max in-degree $M$, optional knowledge $K$(DIR-EDG, UND-EDG, FOR-EDG, REL-TEM v STR-TEM, INI-GRA, VAR-REL, TAR-VAR, REL-BDN v STR-BDN)

**Output:** DAG $G_{max}$ that maximises $S_G = \text{BIC}_{TAR-VAR}(G,D)$ in the search space of DAGs given $K$

**if** $K$(INI-GRA) is not empty **do** $G =$ INI-GRA
    **else do** $G + (K$(DIR-EDG) v randomise direction of $K$(UND-EDG)$) \vDash DAG, M$
**do** $S_G = \text{BIC}_{TAR-VAR}(G,D)$, $S_{\max} = S_G$ and $G_{\max} = G$
**while** $S_{\max}$ increases **do**
    **for** every arc addition, reversal and removal in $G_{\max}$, bounded by CPSs, **do**
        **if** $G_{\max} \vDash DAG, M, K$(DIR-EDG, UND-EDG, FOR-EDG, REL-TEM v STR-TEM) **do**
            $S_{G_n} = \text{BIC}_{TAR-VAR}(G_n, D)$
        **if** $S_{G_n} > S_{\max}$ and $S_{G_n} > S_G$ **do** $G = G_n$, and $G_{\max} = G$
    **end for**
    **if** $S_G > S_{max}$ **do** $S_{max} = S_G$ and $G_{max} = G$
**end while**

// Process for constraint VAR-REL; i.e., variables are relevant
**if** $K$(VAR-REL) is not empty **do**
    **while** independent subgraphs or unconnected variables in $G > 1$ **do**
        HC and return $G_n$ with an extra arc that minimally decreases $\text{BIC}_{TAR-VAR}(G,D)$
        $G = G_n$
    **end while**

// Process for constraints REL-BDN v STR-BDN; i.e., Bayesian Decision Networks (BDNs)
**if** $K$(REL-BDN v STR-BDN) is not empty **do**
    convert learnt graph to a BDN graph
    **if** $K$(STR-BDN) is not empty **do**
        **while** a Decision has no child v a Utility has no parent in $G$ **do**
            HC and return $G_n$ with an extra arc for a Decision/Utility that minimally decreases $\text{BIC}_{TAR-VAR}(G,D)$
            $G = G_n$
        **end while**

**Algorithm 2:** TABU algorithm with additional pseudocode needed to account for the knowledge approaches highlighted in red font colour.

**Input:** dataset $D$, empty graph $G$, objective function $\text{BIC}_{TAR-VAR}(G, D)$, max in-degree $M$, optional knowledge $K$(DIR-EDG, UND-EDG, FOR-EDG, REL-TEM ∨ STR-TEM, INI-GRA, TAR-VAR, REL-BDN ∨ STR-BDN)
**Output:** DAG $G_{max}$ that maximises $S_G = \text{BIC}_{TAR-VAR}(G, D)$ in the search space of DAGs given $K$

**1:** **if** $K$(INI-GRA) is not empty **do** $G =$ INI-GRA
**2:** **else do** $G + (K$(DIR-EDG) ∨ randomise direction of $K$(UND-EDG)) $\vDash DAG, M$
**3:** **do** HC (Algorithm 1) on $G$ and get $S_{\max}$ and $G_{max}$
**4:** **while** $S_{\max}$ increases *or* for up to $|D|(|D| - 1)$ times **do**
**5:** **for** each arc addition, reversal and removal in $G_{max}$, bounded by CPSs, that produces $G_n \vDash DAG$ **do**
**6:** find $G_n$ that minimally decreases $S_{\max}$ where $G_n \vDash DAG, M, K$(DIR-EDG, UND-EDG, FOR-EDG, REL-TEM ∨ STR-TEM)
**7:** **do** HC (Algorithm 1) on $G_n$ and get $S_{\text{n,max}}$ and $G_{n,max}$
**8:** **do** $S_{\text{n,max}} = \text{BIC}_{TAR-VAR}(G_{nn}, D)$
**9:** **if** $S_{\text{n,max}} > S_{\max}$ **do** $S_{\max} = S_{\text{n,max}}$ and $G_{max} = G_{n,max}$
**10:** **else do** remove $G_n$ from the list of neighbours of $G_{max}$
**11:** **end for**
**12:** **end while**

// Process for constraint $VAR - REL$; i.e., variables are relevant
**13:** **if** $K$(VAR-REL) is not empty **do**
**14:** **while** independent subgraphs or unconnected variables in $G > 1$ **do**
**15:** HC and return $G_n$ with an extra arc that minimally decreases $\text{BIC}_{TAR-VAR}(G, D)$
**16:** $G = G_n$
**17:** **end while**

// Process for constraints REL-BDN ∨ STR-BDN; i.e., Bayesian Decision Networks (BDNs)
**18:** **if** $K$(REL-BDN ∨ STR-BDN) is not empty **do**
**19:** convert learnt graph to a BDN graph
**20:** **if** $K$(STR-BDN) is not empty **do**
**21:** **while** a Decision has no child ∨ a Utility has no parent in $G$ **do**
**22:** HC and return $G_n$ with an extra arc for a Decision/Utility that minimally decreases $\text{BIC}_{TAR-VAR}(G, D)$
**23:** $G = G_n$
**24:** **end while**

**Algorithm 3:** Saiyan Hybrid (SaiyanH) algorithm with additional pseudocode needed to account for the knowledge approaches highlighted in red font colour. The MMD score and EMSG graph are described in Appendix A.

**Input:** dataset $D$, variable set $V$, empty graph $G$, objective function $\text{BIC}_{TAR-VAR}(G, D)$, max in-degree $M$, optional knowledge $K$(DIR-EDG, UND-EDG, FOR-EDG, REL-TEM ∨ STR-TEM, INI-GRA, TAR-VAR, REL-BDN ∨ STR-BDN)
**Output:** DAG $G_{max}$ that maximises $S_G = \text{BIC}_{TAR-VAR}(G, D)$ in the search space of DAGs given $K$

// Phase 1: Associational learning
**1:** Get MMD score set for all possible edges
set max MMD score for edges in $K$(DIR-EDG, UND-EDG, INI-GRA)
set min MMD score for edges in $K$(FOR-EDG, REL-TEM ∨ STR-TEM)
**2:** Produce EMSG given MMD

// Phase 2: Constraint-based learning
**3:** Get $MMD_c$ for all conditional dependency tests over $V$, excluding edges $K$(DIR-EDG, INI-GRA, FOR-EDG)
**4:** Get classification set $C$ (cond. dependence/independence/insignificance) over $MMD_c$
**5:** Orientate edges given $K$(DIR-EDG, INI-GRA) and $C$, where EMSG $\vDash DAG, M, K$(REL-TEM ∨ STR-TEM)
**6:** **for** up to two times **do**
**7:** orientate edges given $\text{BIC}_{TAR-VAR}(EMSG, D)$ where EMSG $\vDash DAG, M, K$(REL-TEM ∨ STR-TEM)
**8:** skip unoriented edges when both orientations produce EMSG $\nvDash DAG, M, K$(REL-TEM ∨ STR-TEM)
**9:** orientate edges given $do$-calculus where EMSG $\vDash DAG, M, K$(REL-TEM ∨ STR-TEM)
**10:** skip unoriented edges when both orientations produce EMSG $\nvDash DAG, M, K$(REL-TEM ∨ STR-TEM)
**11:** **end for**

// Phase 3: Score-based learning
Do TABU (Algorithm 2) given $D$, $EMSG$, $\text{BIC}_{TAR-VAR}(G,D)$, $M$, $K$(DIR-EDG, UND-EDG, FOR-EDG, REL-TEM v STR-TEM, INI-GRA, TAR-VAR, and constrain the search space of graphs to those that do not contain independent subgraphs or independent variables

// Process for constraints REL-BDN v STR-BDN; i.e., Bayesian Decision Networks (BDNs)
**if** $K$(REL-BDN v STR-BDN) is not empty **do**
  convert learnt graph to a BDN graph
  **if** $K$(STR-BDN) is not empty **then**
    **while** a Decision has no child v a Utility has no parent in $G$ **do**
      HC and return $G_n$ with an extra arc for a Decision/Utility that minimally decreases $\text{BIC}_{TAR-VAR}(G,D)$
      $G = G_n$
    **end while**

---

**Algorithm 4:** Model-Averaging Hill-Climbing (MAHC) algorithm with additional pseudocode needed to account for the knowledge approaches highlighted in red font colour.

---

**Input:** dataset $D$, empty graph $G$, objective function $\text{BIC}_{TAR-VAR}(G,D)$, max in-degree for pruning phase $M_P$, max in-degree for structure learning phase $M_S$, optional knowledge $K$(DIR-EDG, UND-EDG, FOR-EDG, REL-TEM v STR-TEM, INI-GRA, VAR-REL, TAR-VAR, REL-BDN v STR-BDN).

**Output:** DAG $G_{max}$ that maximises the average set of scores $S_G = \overline{S(G,G_n)} = \overline{\text{BIC}_{TAR-VAR}(G,G_n,D)}$, where $G_n$ represents all the valid neighbours of $G$, in the search space of DAGs given $K$.

**if** $K$(INI-GRA) is not empty **do** $G =$ INI-GRA
  **else do** $G + (K$(DIR-EDG) v randomise direction of $K$(UND-EDG)) $\vDash DAG, M$
Pre-process CPSs $\vDash M_P$, and obtain edge set $F$ containing edges forbidden/pruned off
**do** $S_G = \overline{S(G,G_n)}$, $S_{\max} = S_G$ and $G_{\max} = G$
**while** $S_{\max}$ increases **do**
  **for** every neighbour $G_n$ of $G_{max}$ where $G_n \vDash DAG, M_S, F, K$(DIR-EDG, UND-EDG, FOR-EDG, REL-TEM v STR-TEM) **do**
    add $\text{BIC}_{TAR-VAR}(G_n,D)$ to a new set of objective scores $\boldsymbol{S}$
    **for** every $G_{nn}$ of $G_n$ where $G_{nn} \vDash DAG, M_S, F, K$(DIR-EDG, UND-EDG, FOR-EDG, REL-TEM v STR-TEM) **do**
      add $\text{BIC}_{TAR-VAR}(G_{nn},D)$ to the set of objective scores $\boldsymbol{S}$
    **end for**
    **do** $S_{G_n} = \overline{\boldsymbol{S}}$
    **if** $S_{G_n} > S_G$ **do** $S_G = S_{G_n}$ and $G = G_n$
  **end for**
  **if** $S_G > S_{\max}$, **do** $S_{\max} = S_G$ and $G_{\max} = G$
**end while**

// Process for constraint VAR-REL; i.e., variables are relevant
**if** $K$(VAR-REL) is not empty **do**
  **while** independent subgraphs or unconnected variables in $G >$ 1 **do**
    HC and return $G_n$ with an extra arc that minimally decreases $\text{BIC}_{TAR-VAR}(G,D)$
    $G = G_n$
  **end while**

// Process for constraints REL-BDN v STR-BDN; i.e., Bayesian Decision Networks (BDNs)
**if** $K$(REL-BDN v STR-BDN) is not empty **do**
  convert learnt graph to a BDN graph
  **if** $K$(STR-BDN) is not empty **do**
    **while** a Decision has no child v a Utility has no parent in $G$ **do**
      HC and return $G_n$ with an extra arc for a Decision/Utility that minimally decreases $\text{BIC}_{TAR-VAR}(G,D)$
      $G = G_n$
    **end while**

---

**Algorithm 5:** Greedy Equivalence Search (GES) algorithm applied to the DAG-space with additional pseudocode needed to account for the knowledge approaches highlighted in red font colour.

**Input:** dataset $D$, empty graph $G$, objective function $\text{BIC}_{TAR-VAR}(G, D)$, max in-degree $M$, optional knowledge $K$(DIR-EDG, UND-EDG, FOR-EDG, REL-TEM v STR-TEM, INI-GRA, VAR-REL, TAR-VAR, REL-BDN v STR-BDN)
**Output:** DAG $G_{max}$ that maximises $S_G = \text{BIC}_{TAR-VAR}(G, D)$ in the search space of DAGs given $K$

**if** $K$(INI-GRA) is not empty **do** $G =$ INI-GRA
  **else do** $G + (K(\text{DIR-EDG}) \vee \text{randomise direction of } K(\text{UND-EDG})) \vDash DAG, M$
**do** $S_G = \text{BIC}_{TAR-VAR}(G, D)$, $S_{\max} = S_G$ and $G_{\max} = G$
**while** $S_{\max}$ increases **do**
  **for** every arc addition in $G_{\max}$, bounded by CPSs, **do**
    **if** $G_{\max} \vDash DAG, M, K$(DIR-EDG, UND-EDG, FOR-EDG, REL-TEM v STR-TEM) **do**
      $S_{G_n} = \text{BIC}_{TAR-VAR}(G_n, D)$
    **if** $S_{G_n} > S_{\max}$ and $S_{G_n} > S_G$ **do** $G = G_n$, and $G_{\max} = G$
  **end for**
  **if** $S_G > S_{max}$ **do** $S_{max} = S_G$ and $G_{max} = G$
**end while**

**while** $S_{\max}$ increases **do**
  **for** every arc removal in $G_{\max}$ **do**
    **if** $G_{\max} \vDash DAG, M, K$(DIR-EDG, UND-EDG) **do**
      $S_{G_n} = \text{BIC}_{TAR-VAR}(G_n, D)$
      **if** $S_{G_n} > S_{\max}$ and $S_{G_n} > S_G$ **do** $G = G_n$, and $G_{\max} = G$
  **end for**
  **if** $S_G > S_{max}$ **do** $S_{max} = S_G$ and $G_{max} = G$
**end while**

// Process for constraint VAR-REL; i.e., variables are relevant
**if** $K$(VAR-REL) is not empty **do**
  **while** independent subgraphs or unconnected variables in $G > 1$ **do**
    HC and return $G_n$ with an extra arc that minimally decreases $\text{BIC}_{TAR-VAR}(G, D)$
    $G = G_n$
  **end while**

## 5. Simulation and experiments

The simulation and experiments cover a range of different possible situations one might encounter in practice. In summary, this section describes how each of the 10 knowledge approaches is applied, with varying levels of knowledge constraints, to each of the five algorithms described in Section 4. We consider graphical models that come from real case studies that vary in size and complexity, and use those models to generate synthetic data that vary in sample size under the assumption that the effect of knowledge might depend on the size of the available data.

We evaluate the impact of the first seven knowledge approaches enumerated in Table 1 empirically, determined by how they modify the learnt graph in relation to the ground truth graph. Each approach is evaluated independently, and the evaluation is based on constraints randomly sampled from the ground truth graph. The remaining three approaches TAR-VAR, REL-BDN and STR-BDN represent modelling preferences not necessarily assumed to improve causal structure, and because of this, it would be unreasonable to measure their usefulness in terms of graphical accuracy.

While continuous data has some advantages in terms of learning efficiency, our focus in this paper will be on discrete data that makes no assumptions about the underlying distributions that generated the data, such as an assumption of normality. This is because real data sets are unlikely to be normally distributed, and typically capture many categorical variables. Moreover, all the networks considered in this study come from actual studies, and all the networks are discrete. Starting from the simplest case study, the six networks, which are taken from the Bayesys repository[5], are a) the Asia network, b) the Sports network, c) the Property network, d) the Alarm network, e) the ForMed network, and f) the Pathfinder network.

We consider these six networks suitable for this study for three reasons. Firstly, they represent real-world BNs from a wide range of disciplines taken from the literature, and this makes them appropriate for investigating the impact of knowledge. Secondly, we consider the size of these networks to be representative of most knowledge-based BNs that contain up to hundreds of variables; whereas larger networks containing thousands of variables are impractical for knowledge elicitation. Thirdly, the six networks offer a good range of old and new, as well as simple and complex, BNs that come from different domains. The properties of the six networks are depicted in Table 4, and show that the complexity of the networks ranges between 8 and 109 nodes, 8 and 195 arcs, 2 and 3.58 average degree, 2 and 6 maximum in-degree, and 18 and 71,890 free parameters that are used to approximate the dimensionality of a BN.

**Table 4.** The properties of the six networks.

| Case study | Nodes | Arcs | Average degree | Maximum In-degree | Free parameters |
|---|---|---|---|---|---|
| Asia | 8 | 8 | 2.00 | 2 | 18 |
| Alarm | 37 | 46 | 2.49 | 4 | 509 |
| Pathfinder | 109 | 195 | 3.58 | 5 | 71,890 |
| Sports | 9 | 15 | 3.33 | 2 | 1,049 |
| ForMed | 88 | 138 | 3.14 | 6 | 912 |
| Property | 27 | 31 | 2.30 | 3 | 3,056 |

The experiments for the first seven knowledge approaches are based on synthetic data generated from the six BNs described above. All the six networks come with pre-defined parameters. Specifically, the networks Asia, Alarm and Pathfinder are based on knowledge-based parameters as defined in the relevant study that published the network, the parameters of the networks ForMed and Sports are based on real data, whereas the parameters of network Property were determined by clearly defined rules and regulating protocols involving property tax and other human-made housing market policies.

Synthetic data were sampled from these networks. Five sample sizes are considered for each network, ranging from 100 to 1 million samples; i.e., $10^2$, $10^3$, $10^4$, $10^5$, and $10^6$. The different sample sizes enable us to investigate the impact of knowledge, on networks of varying complexity, given both limited and big data. Overall, the experiments associated with the first seven approaches are based on results obtained from learning 3,245 graphs that required a total structure learning runtime of 44.89 days. Specifically, each of the ten knowledge approaches was applied to each of the five algorithms, with different rates of constraints as specified in Table 5, and in combination with the six case studies and five different sample sizes per case study.

**Table 5.** An example of the number of constraints simulated for each knowledge approach and rate combination, based on the Alarm network.

| Approach ID | Knowledge approach | # of edge constraints applied to: 5% | 10% | 20% | 50% | 100% | Varies | Variables constrained: 5% | 10% | 20% | 50% | 100% | Varies |
|---|---|---|---|---|---|---|---|---|---|---|---|---|---|
| DIR-EDG | Directed edge | 2 | 5 | 9 | 23 | - | - | - | - | - | - | - | - |
| UND-EDG | Undirected edge | 2 | 5 | 9 | 23 | - | - | - | - | - | - | - | - |
| FOR-EDG | Forbidden edge | 2 | 5 | 9 | 23 | - | - | - | - | - | - | - | - |
| REL-TEM | Relaxed incomplete temporal order | - | - | - | - | - | - | 2 | 4 | 7 | 19 | - | - |
| STR-TEM | Strict incomplete temporal order | - | - | - | - | - | - | 2 | 4 | 7 | 19 | - | - |
| INI-GRA | Initial graph | - | - | 9 | 23 | - | - | - | - | - | - | - | - |
| VAR-REL | Variables are relevant | - | - | - | - | - | ≥0 | - | - | - | - | 37 | - |
| TAR-VAR | Target variable/s | - | - | - | - | - | ≥0 | - | - | - | - | - | ≥0 |
| REL-BDN | Relaxed BDNs | - | - | - | - | - | - | - | - | - | - | - | - |
| STR-BDN | Strict BDNs | - | - | - | - | - | ≥0 | - | - | - | - | - | ≥0 |

[5] http://bayesian-ai.eecs.qmul.ac.uk/bayesys/

Table 5 presents, as an example, the number of constraints simulated for the Alarm network, for each knowledge approach and corresponding rate of constraints combination. The rates of constraint selected are those deemed to be reasonable for each approach. For example, we assume that hard knowledge for approaches such as DIR-EDG is more likely to range between 5% to 50% of edges relative to those present in the true graph. Conversely, if a user decides to incorporate soft knowledge via INI-GRA, then it makes more sense that the rate of constraints would be higher and possibly ranging from 20% to 50%. Therefore, comparisons about the impact of each of these approaches are measured overall, and across all rates of constraints sampled.

The constraints are randomly sampled from the ground truth graph. For example, the number of constraints in DIR-EDG and UND-EDG is determined relative to the number of edges in the true graph. Given that the Alarm network contains 46 edges, 23 of those edges are randomly selected as constraints when the rate is set to 50% (note the same edges are selected for both DIR-EDG and UND-EDG). For lower constraint rates, we sample part of those randomised at the 50% rate; e.g., at rate 20% we select the first nine edges of the 23 edges determined at 50% rate. While it can be argued that the number of forbidden edges (approach FOR-EDG) should be a percentage of the independencies (i.e., edges absent) in the ground truth graph, in an analogous approach to that used for edges present, we decided to follow the same quantities of constraints as in DIR-EDG and UND-EDG for consistency, but also because there are

$$\frac{|V|(|V|-1)}{2} - a \tag{3}$$

missing edges compared to the completely connected graph in a DAG, where $a$ is the number of edges and $V$ the variable set. For example, the Alarm network contains 620 missing edges, and it would represent an unrealistic scenario to sample 50% of those independencies as constraints.

Temporal information (i.e., approaches REL-TEM and STR-TEM) is trickier to sample and there are multiple ways this can be done. We followed a simple process where the number of tiers is determined by the DAG structure in which child nodes are always ordered under their parents, as illustrated in Fig 3 with reference to the Sports network. We then sample a percentage of the variables (as opposed to edges) and assign them to a tier as illustrated in Fig 3. Since temporal constraints require at least two variables as input, smaller networks such as Asia and Sports cannot be tested at lower rates of temporal constraints. This is because selecting just two, out of the eight or nine nodes available in the Asia and Sports networks respectively, corresponds to a rate higher than 5% and 10%.

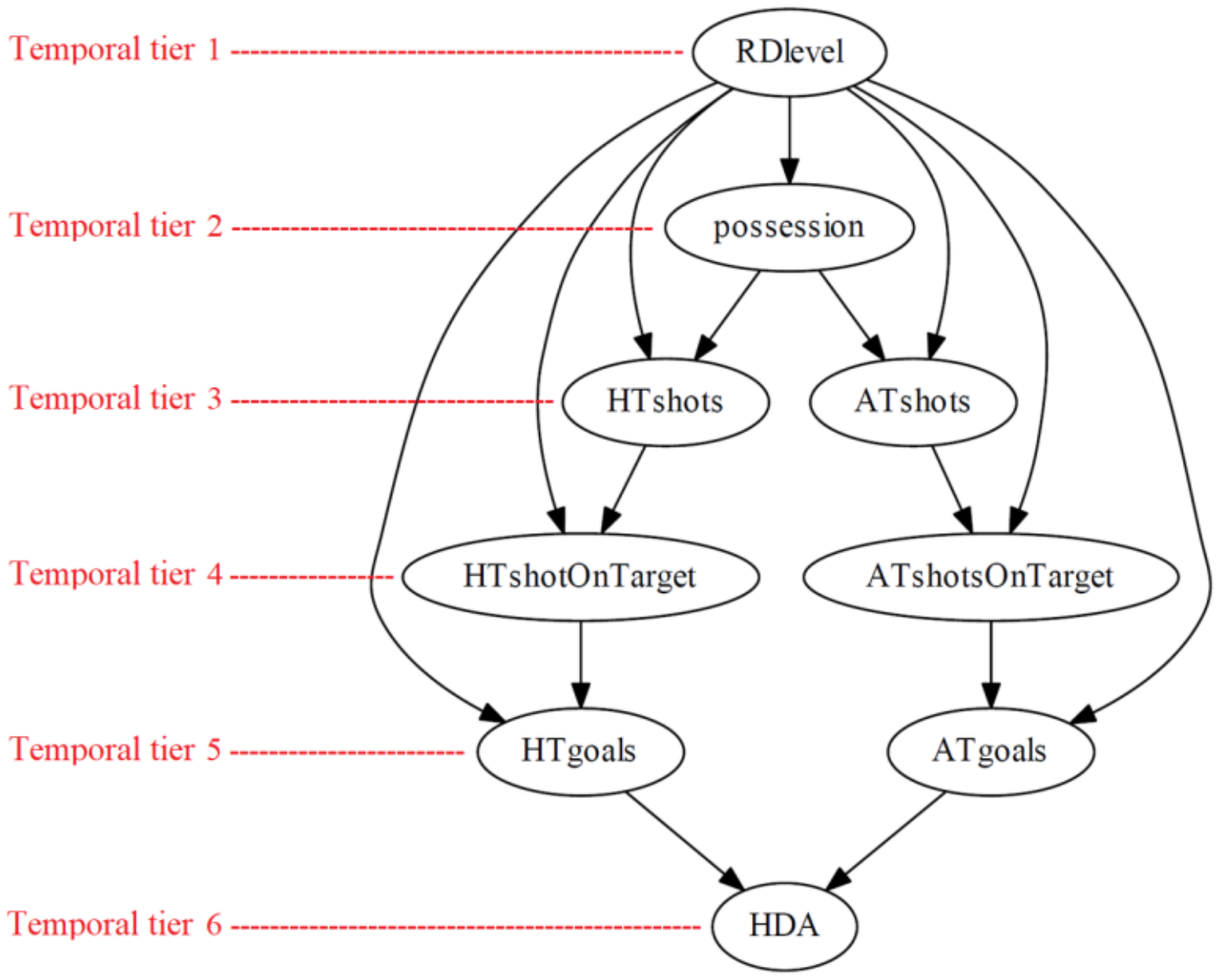

**Fig 3.** The ground truth graph of the Sports network with node rows corresponding to temporal tiers. The graph is generated in Bayesys which makes use of the Graphviz visualisation system [24].

For INI-GRA we assume the rates of 20% and 50% (refer to Table 5) on the basis most users who select this approach would be intending to guess at least half of the graph. To enable unbiased comparisons between the different knowledge approaches, the experiments at the 50% rate reuse the same edges selected for the same rates in DIR-EDG and UND-EDG. Note that unlike the other approaches examined, INI-GRA represents a soft constraint that affects structure learning by specifying the starting position in the search space of graphs, and there is no guarantee it will preserve the input knowledge edge-set in the learnt output, unlike the DIR-EDG and UND-EDG approaches.

The remaining knowledge approaches operate somewhat differently than the rest. Specifically, approach VAR-REL involves no randomisation since it forces the algorithm to return a graph that contains no independent subgraphs or nodes. As a result, VAR-REL can be viewed as an approach that may constrain all of the variables, as indicated in Table 5. Moreover, it also involves an unknown number of edges required to connect all the potentially independent subgraphs and nodes, indicated as '≥0' under edge rate *Varies* in Table 5. Note this approach makes no difference to the output of SaiyanH, since it incorporates this approach by design.

Further to what has been discussed at the beginning of this section, the impact of approaches TAR-VAR, REL-BDN, and STR-BDN is not measured using synthetic data due to the nature of these approaches. Specifically, approach TAR-VAR is applied to two real health datasets where the target variables are driven by author needs, as indicated in the relevant publications [16, 29] which we discuss in more detail in Section 6, whereas the practicality of approaches REL-BDN and STR-BDN is shown via illustration of network modification.

## 6. Results and discussion

The discussion of the results focuses on measuring the impact each knowledge approach has on structure learning. We investigate the impact of knowledge using different measures that tell us something about structure learning accuracy and model complexity, to answer the research questions discussed in Section 1. On this basis, the results are presented with a focus on the knowledge approaches and not with a focus on the algorithms, since the approaches can be implemented in almost any algorithm. While each algorithm is expected to benefit differently, the general expectation is that the more accurate an algorithm is with a particular dataset, the less it is expected to benefit from knowledge.

We divide this section into three subsections. Subsection 6.1 presents the results from synthetic empirical evaluation of the first seven approaches described in Table 1, subsection 6.2 illustrates the usefulness of TAR-VAR, and subsection 6.3 the usefulness of REL-BDN and STR-BDN. To better understand the relative impact each knowledge approach has on each algorithm, Table B1 in Appendix B presents how each approach influences the learning performance of each algorithm under different measures.

Various criteria are used to evaluate the first seven approaches. To begin with, graphical accuracy is determined by three scoring metrics that quantify the accuracy of a learnt DAG with respect to the ground truth DAG. The first two metrics are the classic F1 and SHD scores. The third metric, the Balanced Scoring Function (BSF), is relatively new [17] and addresses the bias in graphical score present in other metrics. For example, the SHD score is known to be biased towards the sensitivity of identifying edges versus specificity. The BSF metric eliminates bias by taking into consideration all of the confusion matrix parameters, and balances the score between dependencies present and absent in the true graph. In practice, the BSF score ranges from -1 to 1, where -1 corresponds to the worst possible graph; i.e., the reverse of the true graph that maximises False Positive (FP) and False Negative (FN) edges, and minimises True Positive (TP) and True Negative (TN) edges. Furthermore, 1 to the graph that matches the true graph, and 0 represents the ignorant case of an empty or a fully connected graph. It is calculated as follows:

$$\mathrm{BSF} = 0.5\left(\frac{\mathrm{TP}}{a} + \frac{\mathrm{TN}}{i} - \frac{\mathrm{FP}}{i} - \frac{\mathrm{FN}}{a}\right) \tag{4}$$

where $a$ is the number of edges present and $i$ is the number of edges absent in the true graph, and $i$ is the output of Equation 3; i.e., $i = \frac{|V|(|V|-1)}{2} - a$. Lastly, we assume that arc reversals generate half the penalty of arc deletions and arc additions, on the basis that an arc reversal represents correct dependency with incorrect directionality, and this assumption applies to all the three metrics.

Because structure learning algorithms tend to return a DAG which is consistent with the Complete Partial DAG (CPDAG), which represents a set of Markov equivalent DAGs, it is common to measure the performance of structure learning algorithms in terms of their ability to recover the true CPDAG, rather than the true DAG. This means that while CPDAG-based evaluation considers, for example, $A \rightarrow B \rightarrow C$ to be equivalent to both $A \leftarrow B \leftarrow C$ and $A \leftarrow B \rightarrow C$, a DAG evaluation does not. Because this paper focuses on the impact of knowledge, which we assume to be causal, we choose to measure this impact in terms of DAG structure so that we can differentiate between examples such as the above, since this is part of the purpose of considering knowledge. As we show in Appendix C, however, a CPDAG-based evaluation generates very similar results to the DAG-based evaluation.

In addition to graphical accuracy, we measure the impact of knowledge in terms of model selection, using the BIC as defined in Equations 1 and 2 but excluding the optional term $r$ used for TAR-VAR, where a higher BIC score indicates a better model. We also consider the impact on the number of free parameters as defined in Equation 2, as well as the difference in the number of arcs learnt. Lastly, we also measure the impact on runtime.

### *6.1. Evaluation of the first seven knowledge approaches in Table 1*

Fig 4 presents the average impact, with standard deviation superimposed, each of the first seven approaches had on the graphical accuracy of five algorithms and in terms of F1, BSF and SHD scores, as well as across all six case studies and five sample sizes. The impact is presented in terms of overall, as well as in terms of limited and big data. The motivation here was to investigate the usefulness of the knowledge approaches with different sample sizes relative to the complexity of the network. For example, the sample size of $10^4$ is considered 'big' for the complexity of the Asia network, but 'limited' for the complexity of the Pathfinder network. For the networks Asia, Sports and Property, we assume a sample size of $10^3$ or lower represents limited data, whereas for the larger networks of Alarm, ForMed and Pathfinder we assume $10^4$ or lower represents limited data; sample sizes above these are assumed to represent big data (i.e., sufficient sample size).

The results suggest that the most useful approaches, in terms of improving the graphical accuracy of the learnt graph, are DIR-EDG, UND-EDG and INI-GRA, all of which involve information about edges, and approaches DIR-EDG and INI-GRA also include information about the direction of the edges. Approach DIR-EDG is the one that improves accuracy the most when performance is compared over the same rates of constraint, followed by approach INI-GRA which imposes soft constraints, and approach UND-EDG that entails less information than DIR-EDG and INI-GRA. Interestingly, when the data are limited, DIR-EDG at 50% rate outperforms INI-GRA at 50% rate, although this observation is relaxed or reverses for big data. This might be because when the algorithms explore the search space of DAGs with limited data, they are prone to less accurate model fitting scores and this presumably increases the level of modification the algorithms might perform on the initial graph, thereby the 50% input graph as a less effective soft constraint compared to the 50% rate of directed edges imposed as hard constraints.

On the other hand, the constraints that involve forbidden edges, either directly through approach FOR-EDG or indirectly through the temporal approaches REL-TEM and STR-TEM, appear to be many times less effective compared to constraints that provide information about the presence of edges. This is not surprising since, in the case of temporal constraints, we are selecting relatively few edges to forbid from the relatively large number of edges absent in the ground truth graph. It is important to highlight that these constraints are found to be much more useful with big data rather than with limited data, and this is counterintuitive since knowledge is assumed to be more useful when available data are limited. Still, it can be explained by the fact that structure learning algorithms tend to produce a higher number of edges when trained with big data, and many of those edges could be false positives that these constraints may prohibit. On this basis, the observation that strict temporal order STR-TEM provides only marginal difference in impact relative to REL-TEM is not entirely surprising, since STR-TEM involves only a few more forbidden edges compared to those imposed by REL-TEM.

Lastly, VAR-REL is the only approach that tends to lower the accuracy of the learnt graphs. This result is not unexpected given that when this constraint was first introduced as a feature of SaiyanH, it was recognised that forcing the algorithm to connect all of the data variables implies that the additional forced edges may not be as accurate as those discovered unrestrictedly, and that the benefit of assuming the input variables are dependent comes in the form of practical usefulness by enabling full propagation of evidence [15]. In line with approaches FOR-EDG, REL-TEM and STR-TEM, approach VAR-REL has less detrimental effect with big data than limited data. This can be explained by the fact that less additional, but more accurate, forced edges are likely with big data, compared to the edges identified with limited data. Therefore, VAR-REL exchanges minor graphical accuracy for practical usefulness in the form of full propagation of evidence (not accounting for the accuracy of propagation), and where the risk of a negative effect on graphical accuracy decreases with sample size.

Fig 5 repeats the analysis of Fig 4 and investigates the impact of constraints on the BIC score, the number of free parameters, and the number of arcs learnt. The higher standard deviation, including some inconsistencies observed in the results of UND-EDG could be due to randomisation; i.e., UND-EDG is the only approach that involves randomisation, since the specified undirected edges are directed at random to form a valid initial graph. INI-GRA is the only approach that leads to meaningful improvements in the BIC score, although this improvement occurs only for big data. Moreover, the positive effect decreases at 50% rate of constraints; i.e., when the input graph matches the ground truth. Interestingly, while the impact on BIC appears to be rather marginal overall, it is considerably detrimental in the presence of limited data and further increases with the rate of constraints. This is because BIC is a model selection function that partly expresses how well the learnt distributions fit the data. Because the fitting between the learnt and observed distributions tends to increase with higher sample size, the constraints imposed in the presence of limited data force the learnt distributions to deviate from the observed distributions more strongly. This observation is also supported by the number of free parameters and the number of edges learnt, which both appear to increase faster when constraints are imposed with limited, as opposed to big, data; implying that the forced constraints must have increased the dimensionality penalty $p$ in BIC (refer to Equation 2) faster than they increased the Log-Likelihood score (refer to Equation 1).

Lastly, Fig 6 repeats the previous analyses with reference to runtime. Overall, the results suggest that there is little benefit in runtime by imposing constraints in the process of structure learning. The only approach that appears to meaningfully reduce runtime is INI-GRA at 50% rate, and this is because it helps the algorithms to converge to a local or global maximum faster compared to starting from an empty graph. Overall, the results from runtime can be viewed as counterintuitive on the basis that constraints reduce the search space of graphs, and yet many of the constraints increase rather than decrease runtime. However, while it is true that the search space of graphs is reduced, there is no guarantee that the search space explored will be reduced. This is because the algorithms often explore only a minor proportion of the available search space, and the constraints might set up a tension between what the data indicate and what the constraints are trying to enforce, and this might cause the algorithms to explore different portions of the search space.

Moreover, the temporal approaches REL-TEM and STR-TEM appear to have significant negative repercussions on runtime. This is another result that may appear counterintuitive in the first instance. It is important to clarify that node-ordering algorithms that rely on complete information of the ordering of the variables, such as the K2 algorithm [19] generally reduce the search space of graphs from super-exponential to exponential [42] and hence, tend to reduce runtime[6] considerably. Because the incomplete temporal ordering approaches implemented in this study extend the temporal checks to ancestral relationships for every graph explored, in addition to parental relationships, they increase computational complexity and runtime considerably, something which is avoided by algorithms such as K2 which assume a complete ordering.

[6] The computational complexity of node-ordering algorithms that do not rely on knowledge explore the graphs in different node-ordering spaces and hence, have a higher computational complexity than K2 depending on the number of node-orderings explored (Scanagatta, 2018).

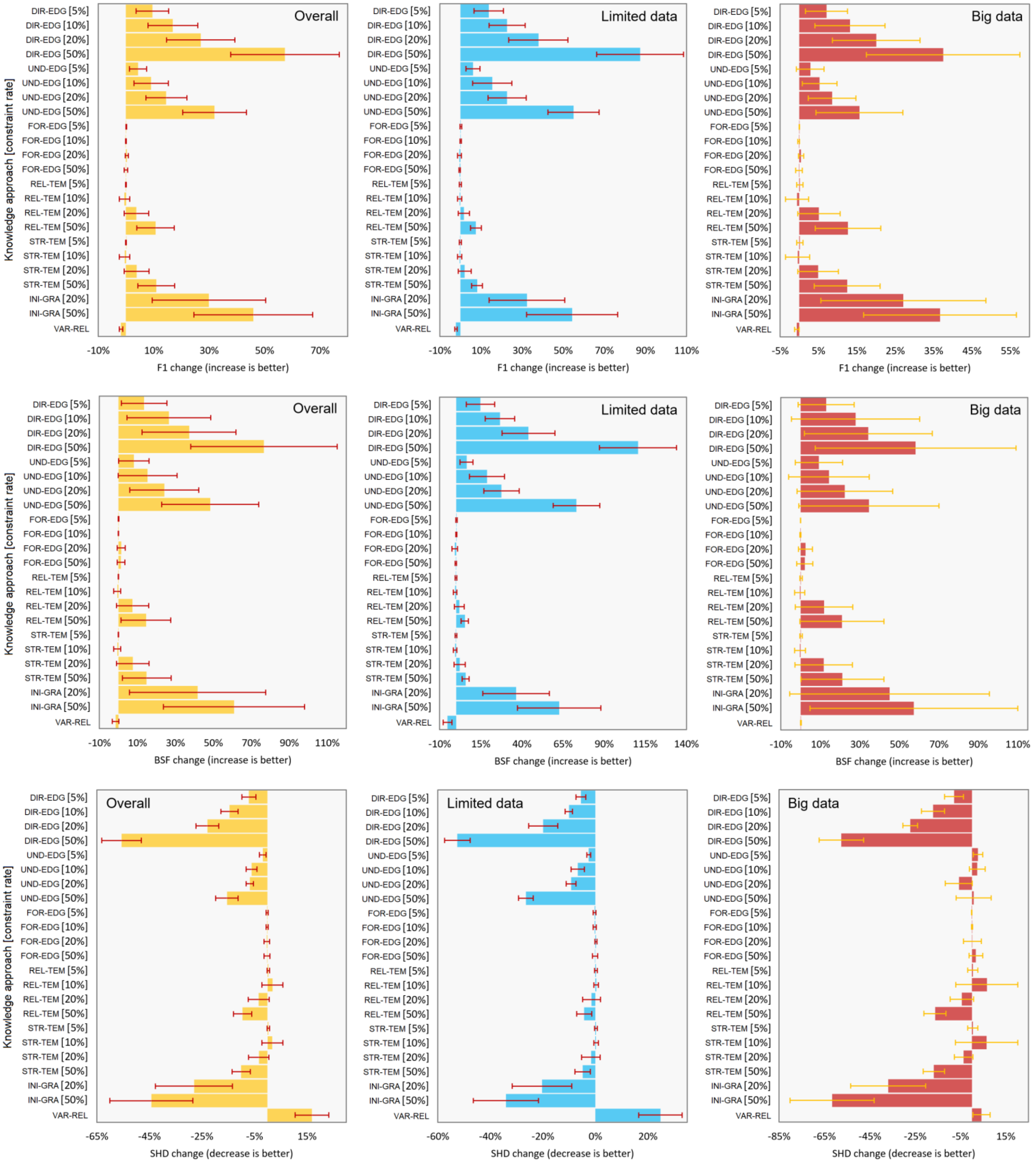

**Fig 4.** The relative impact each knowledge approach has on structure learning performance, and over different rates of constraint, in terms of F1, BSF and SHD scores, where DIR-EDG is directed edges, UND-EDG is undirected edges, FOR-EDG is forbidden edges, REL-TEM is relaxed temporal order, STR-TEM is strict temporal order, INI-GRA is input graph, and VAR-REL is variables are relevant. Error lines represent standard deviation.

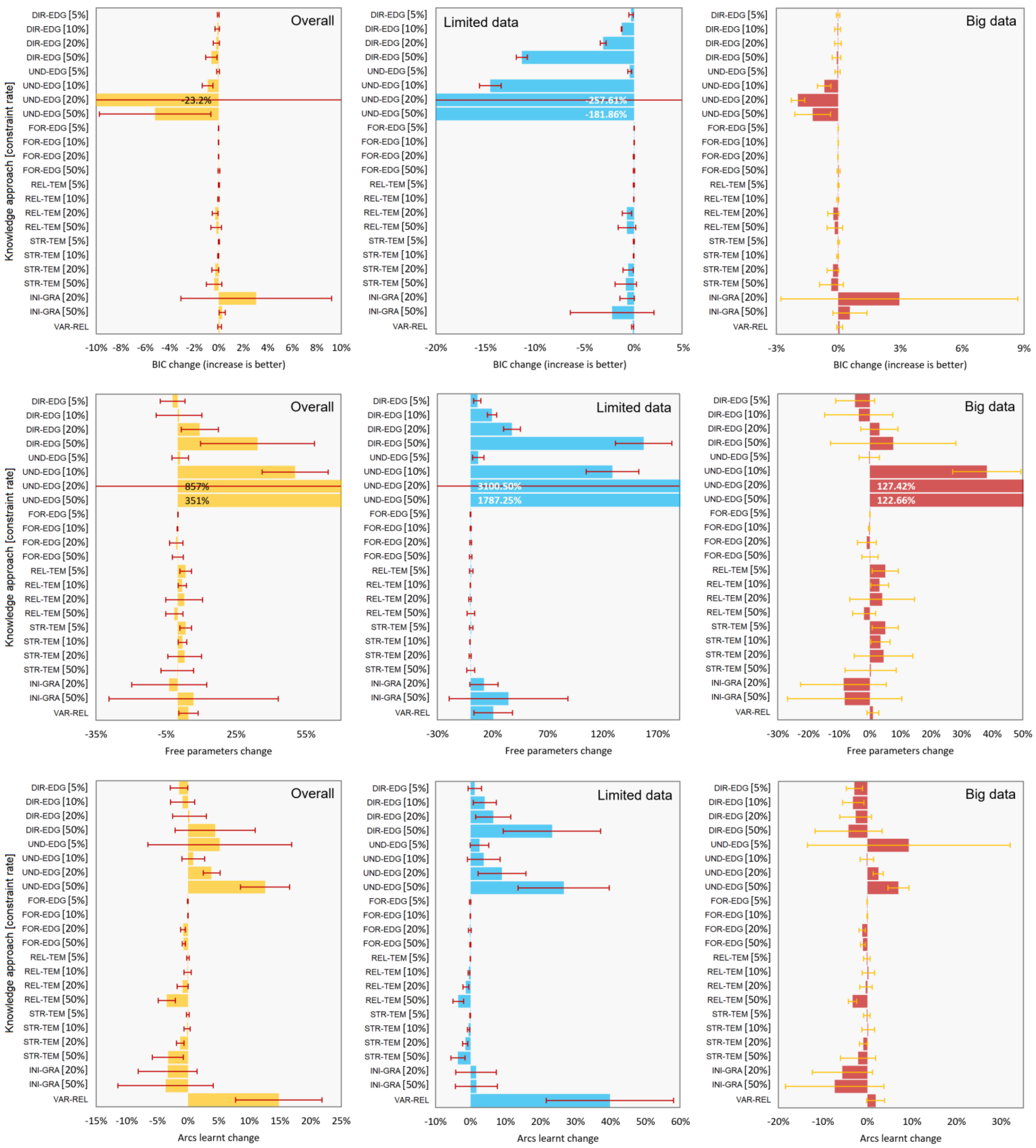

**Fig 5.** The relative impact each knowledge approach has on structure learning performance, and over different rates of constraint, in terms of BIC score, the number of free parameters, and the number of arcs learnt, where DIR-EDG is directed edges, UND-EDG is undirected edges, FOR-EDG is forbidden edges, REL-TEM is relaxed temporal order, STR-TEM is strict temporal order, INI-GRA is input graph, and VAR-REL is variables are relevant. Error lines represent standard deviation.

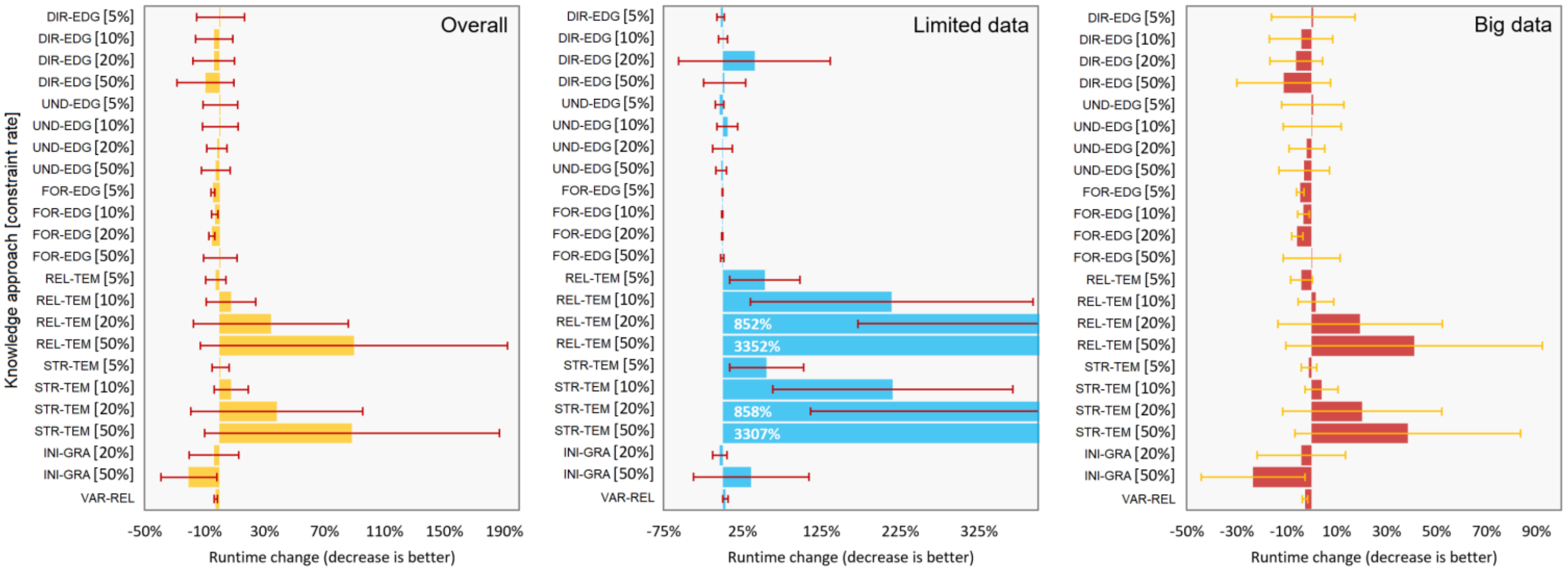

**Fig 6.** The relative impact each knowledge approach has on structure learning runtime, and over different rates of constraint, where DIR-EDG is directed edges, UND-EDG is undirected edges, FOR-EDG is forbidden edges, REL-TEM is relaxed temporal order, STR-TEM is strict temporal order, INI-GRA is input graph, and VAR-REL is variables are relevant. Error lines represent standard deviation.

### 6.2. *Evaluation of knowledge approach* TAR-VAR

Approach TAR-VAR, which involves indicating one or more target variables of interest for which we would like to encourage the algorithm to discover more parents or causes, is evaluated using two real health datasets. These are a) the forensic psychiatry dataset in [16] which captures data about released prisoners during and after release who suffer from serious mental health problems, and b) the demographic and health survey dataset on livelihood factors associated with childhood diarrhoea in [29]. The forensic psychiatry dataset contains 56 variables and a sample size of 953, thereby representing a case study with limited data, whereas the childhood diarrhoea dataset contains 28 variables and a sample size of 259,628, representing a case study with big data.

Both studies reported a variable deemed more important than all other variables. These are a) the variable called *Violence* in the forensic psychiatry dataset which captures information about the risk of a prisoner becoming violent following their release into the community, and b) the variable called *DIA_HadDiarrhoea* in the childhood diarrhoea dataset which captures information about whether a child had diarrhoea in two weeks preceding the survey data collection. Both studies focus on investigating the causal explanations with reference to the target variables. As described in Section 3, approach TAR-VAR enables us to relax the dimensionality penalty for variables of interest to discover a potentially higher number of causes, thereby improving its predictive accuracy and identification of intervention. This can be useful when algorithms produce too sparse a network, often due to limited data.

The dimensionality penalty is relaxed by increasing $r$ as defined in Equation 2, where higher $r$ values encourage the algorithm to explore more complex Conditional Probability Tables (CPTs) for the target variable in terms of the number of free parameters, which in turn makes it more likely for the target variable to contain a higher number of parents in the learnt graph. Fig 7 presents the Markov blankets produced by HC, TABU and GES for target variable *Violence* (the first case study) over different $r$ inputs, including the standard scenario that involves no manipulation of the dimensionality penalty, when $r = 1$. The Markov blanket of node *Violence* represents the node-set that contains all the information one needs to infer *Violence*, and consists of the parents of *Violence*, its children, and the parents of its children. Similarly, Figs 8 and 9 presents the results produced by SaiyanH and MAHC respectively. These illustrations represent a classic example whereby available data are big in terms features and small in terms of sample size, and tend to produce sparse networks with smaller Markov blankets, such as those shown for case ‘standard’ in Figs 7 to 9. Sparse networks are often viewed inadequate for inference and intervention, and this is one of the reasons why practitioners often favour manual graphical construction of BN models applied to real data.

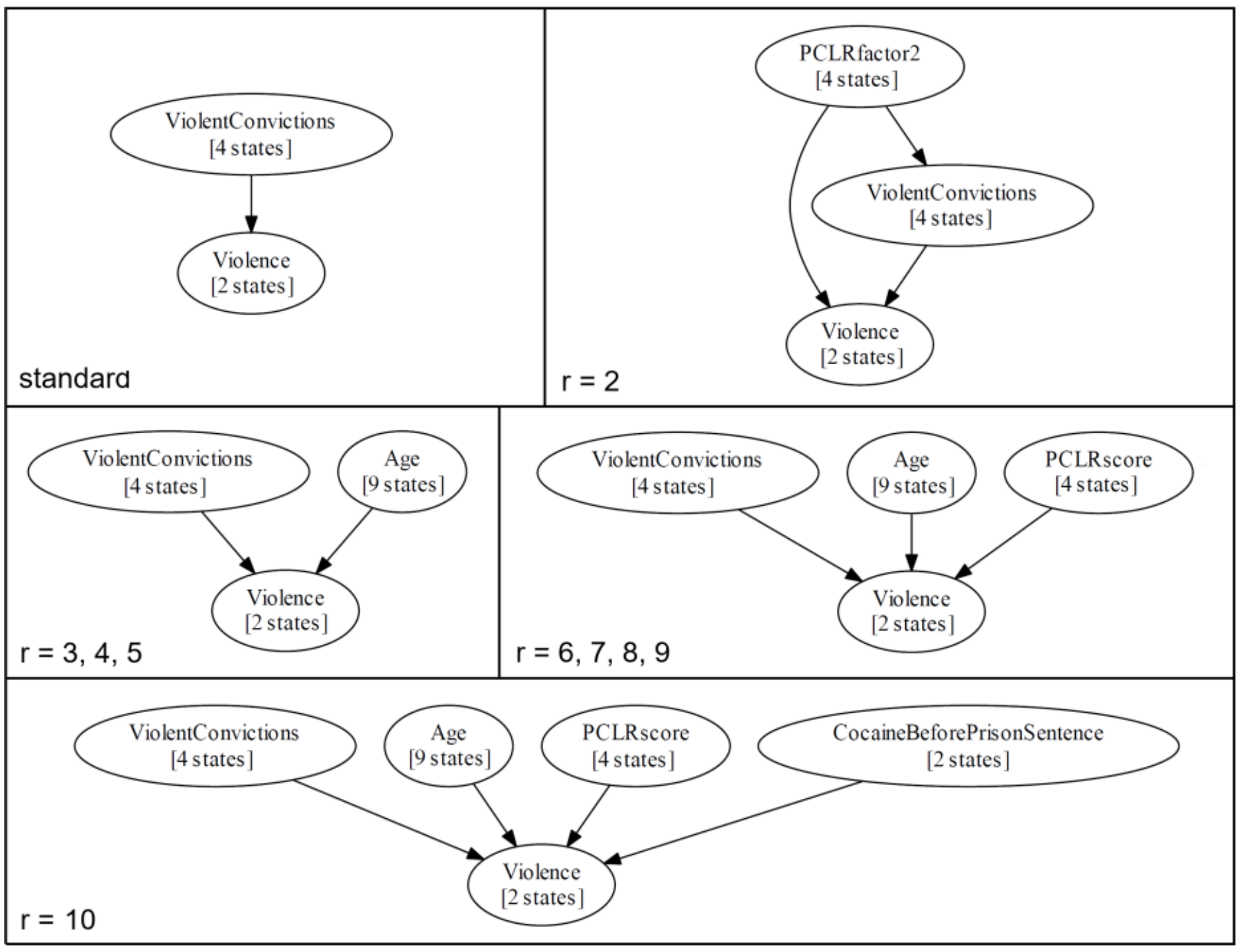

**Fig 7.** The Markov blankets of target variable *Violence* produced by HC, TABU and GES (they produced the same Markov blankets), given the forensic psychiatry data, over different inputs $r_i$ while applying TAR-VAR (refer to Equation 2).

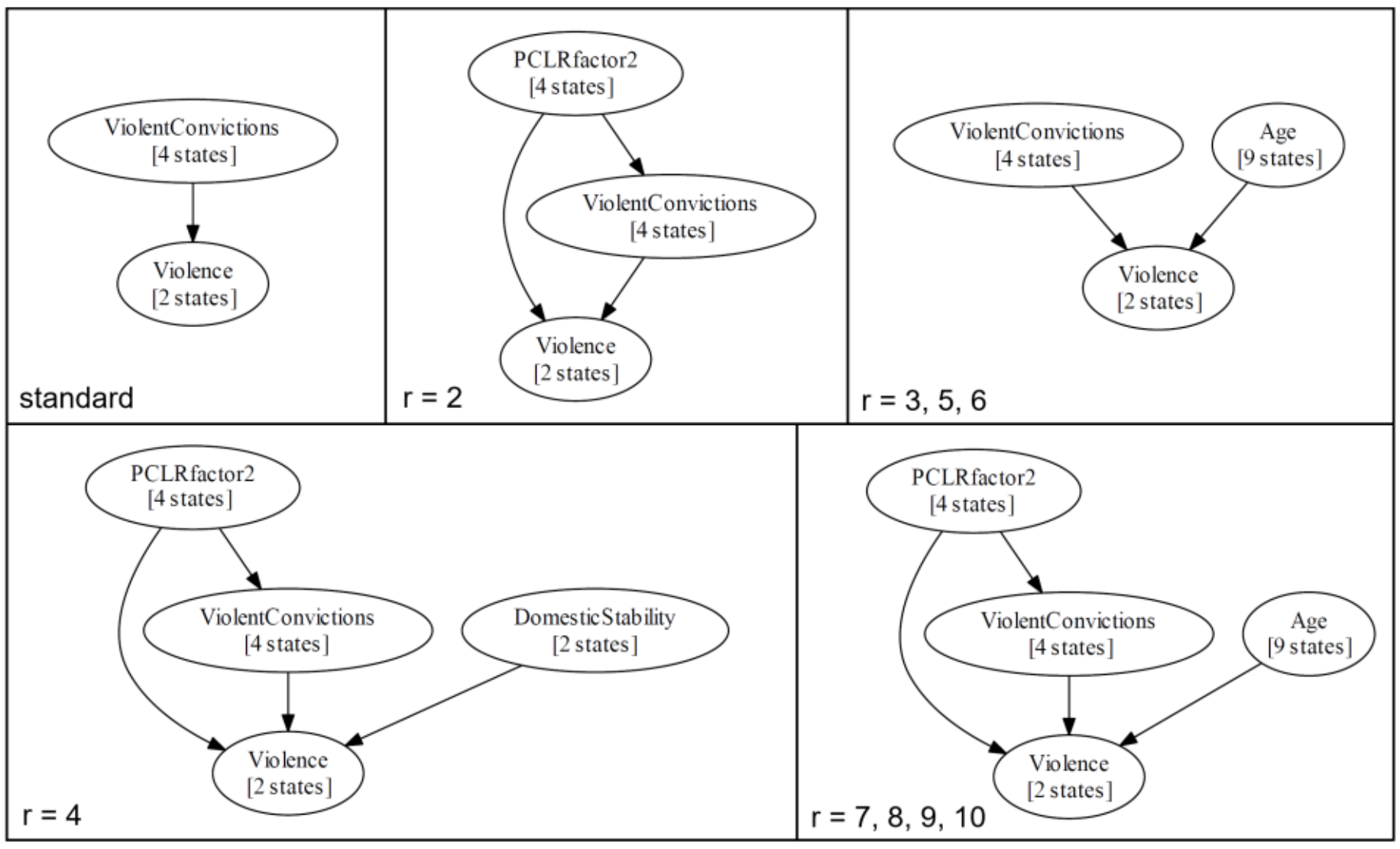

**Fig 8.** The Markov blankets of target variable *Violence* produced by SaiyanH, given the forensic psychiatry data, over different inputs $r_i$ while applying TAR-VAR (refer to Equation 2).

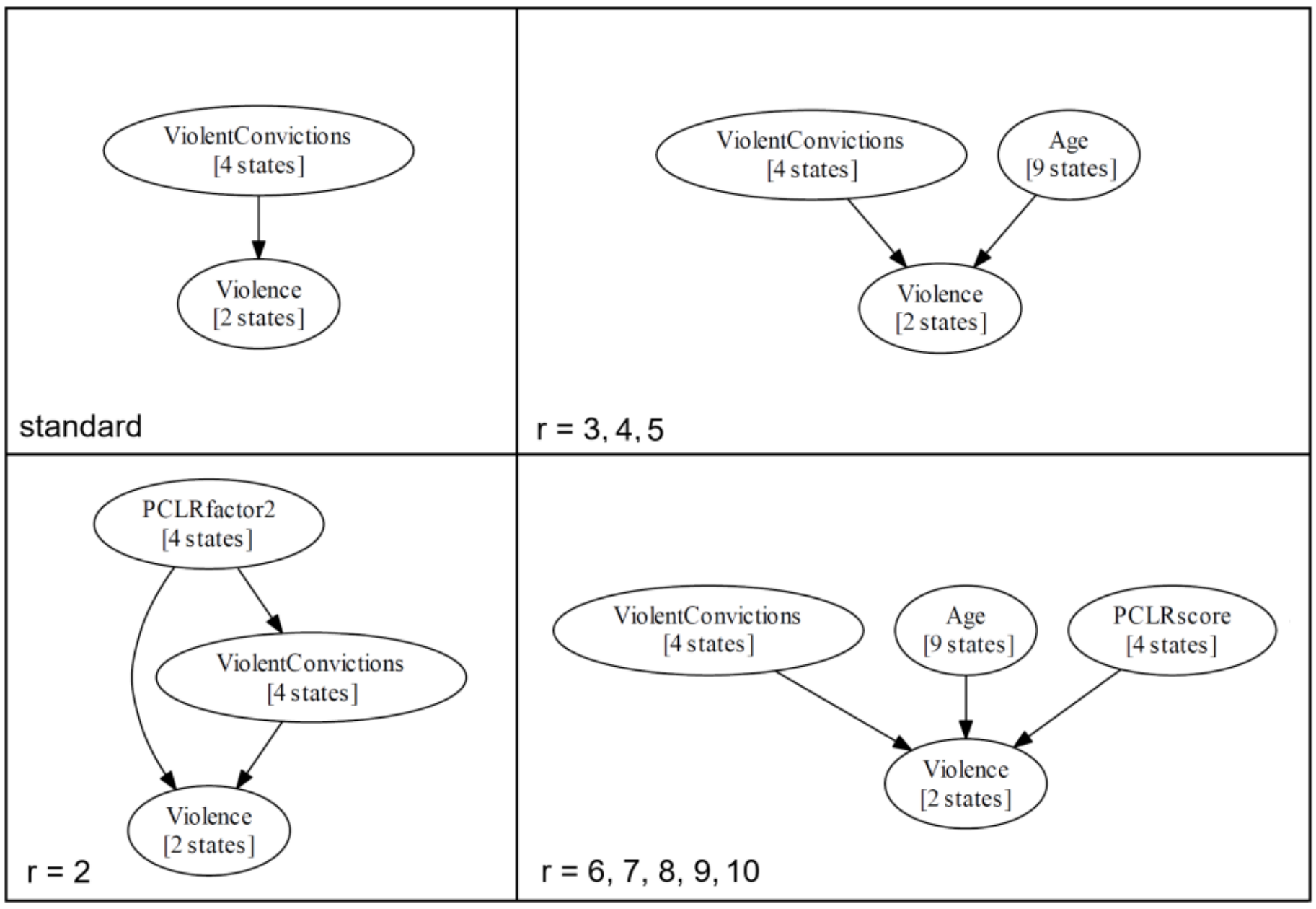

**Fig 9.** The Markov blankets of target variable *Violence* produced by MAHC, given the forensic psychiatry data, over different inputs $r_i$ while applying TAR-VAR (refer to Equation 2).

Table 6 presents information on how the number of free parameters is adjusted over the different $r_i$ inputs, and how they influence other parts of the network, with respect to the Markov blankets shown in Figs 7 to 9 across the five algorithms. The unweighted free parameter $p_u$ represents the number of free parameters when $r = 1$ (i.e., the output is equal to the standard BIC when no variables are targeted as of interest), whereas the weighted $p_w$ represents the adjusted number of free parameters when $r > 1$. For example, the dimensionality penalty of *Violence* is reduced from 16 to 8 when $r$ increases from 1 to 2 (i.e., divided by $r$) in all five algorithms, thereby encouraging structure learning to introduce one more parent of *Violence* as illustrated in Figs 7 to 9. While approach TAR-VAR is unlikely to modify other parts of the graph, since it is applied locally to a decomposable score such as the BIC, it is still possible for modifications other than those affecting the targeted variable to occasionally occur. Out of the 40 instances presented in Table 6, extra network modifications occurred in six of them. Three of those cases involve the TABU algorithm when $r_i = 2, 3, 5$, the fourth case involves the SaiyanH algorithm when $r_i = 4$, and the remaining two cases involve the MAHC algorithms when $r_i = 2, 8$. The additional modifications are minor and can be explained due to variability which arises by manipulating $BIC_{TAR-VAR}$ over different values of $r_i$, and this variability might cause $BIC_{TAR-VAR}$ to optimise at a neighbouring or at a slightly different graph.

Figs 10 to 12, and Table 7 repeat the analysis of Figs 7 to 9 and Table 6, this time with application to the childhood diarrhoea dataset and with *DIA_HadDiarrhoea* set as the target variable. This represents a big data case study since it has a smaller number of variables and much larger sample size than the previous case study. The higher sample size partly explains why the Markov blankets in Figs 10 to 12 are more complex compared to those in Figs 7 to 9. This time, however, applying TAR-VAR on SaiyanH did not produce any additional edges for the target variable, whereas it did for the other three algorithms at a similar rate to the forensic psychiatry dataset. SaiyanH's Markov blanket[7] remained static across the different $r_i$ inputs. This result can also be observed in Table 7 which shows that the unweighted number of free parameters $p_u$ remains static over $r_i$ for SaiyanH. The high sample size of the childhood diarrhoea dataset increases the Log-Likelihood component of the BIC score relative to the complexity component and in turn reduces the importance of $r_i$, thereby making it more difficult for changes in dimensionality penalty to have an impact on the learnt graph relative to the forensic psychiatry dataset.

[7] The Markov Blanket of SaiyanH was "*BF_BreastfedMonths* → *DIA_HadDiarrhoea* ← *GEO_Region*".

The overall results suggest that TAR-VAR can also be useful in the presence of big data. Moreover, we found the influence of $r$ to be well-behaved in the sense that as $r$ increases so do the number of parents of the target variable, as one might hope.

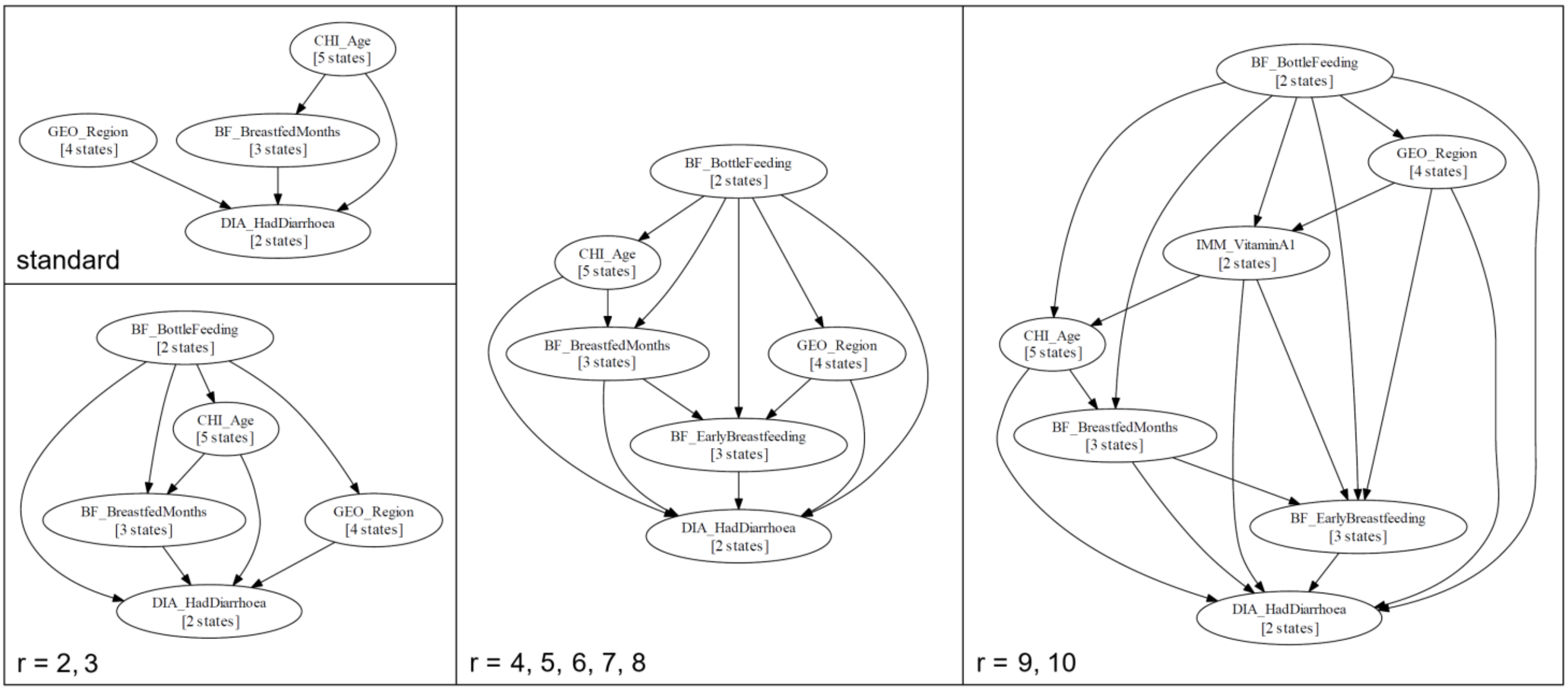

**Fig 10.** The Markov blankets of target variable *DIA_HadDiarrhoea* produced by HC and TABU (they produced the same Markov blankets) given the childhood diarrhoea dataset, over different inputs $r_i$ while applying TAR-VAR (refer to Equation 2).

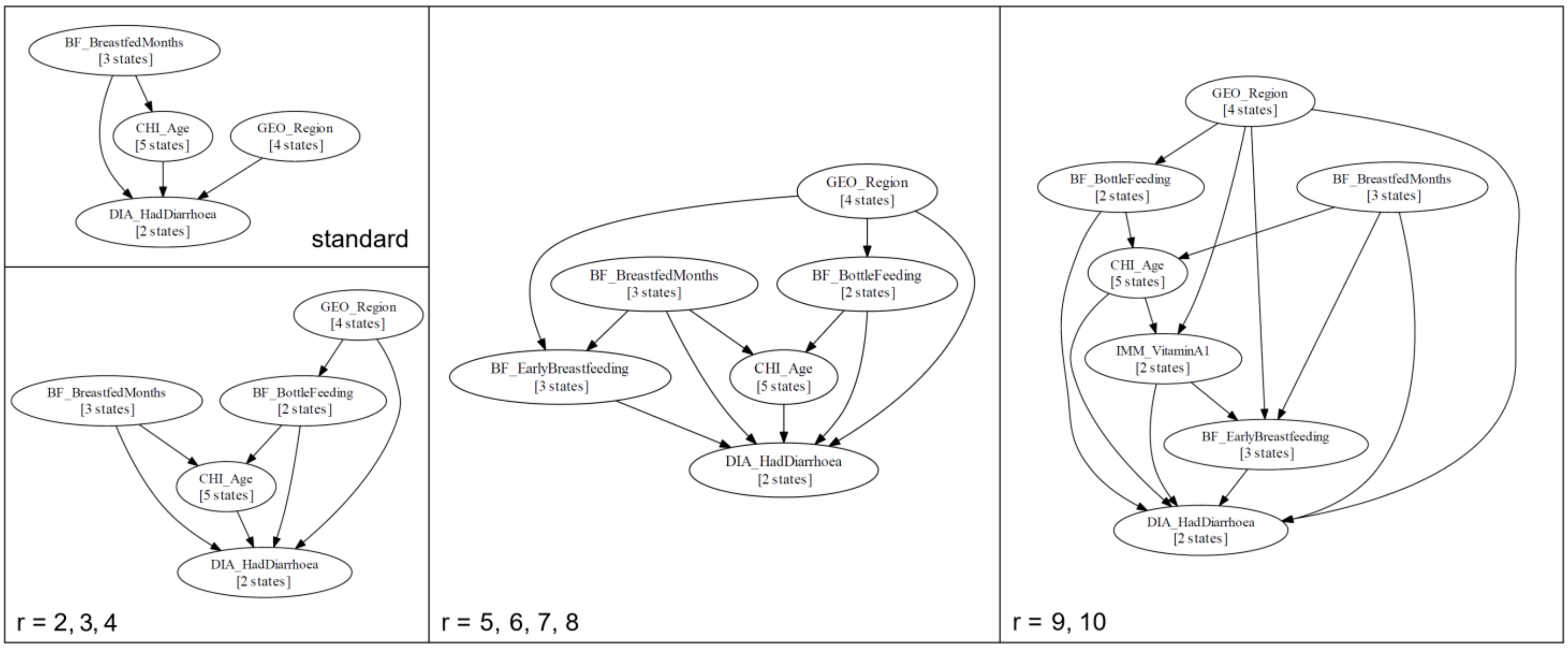

**Fig 11.** The Markov blankets of target variable *DIA_HadDiarrhoea* produced by MAHC given the childhood diarrhoea dataset, over different inputs $r_i$ while applying TAR-VAR (refer to Equation 2).

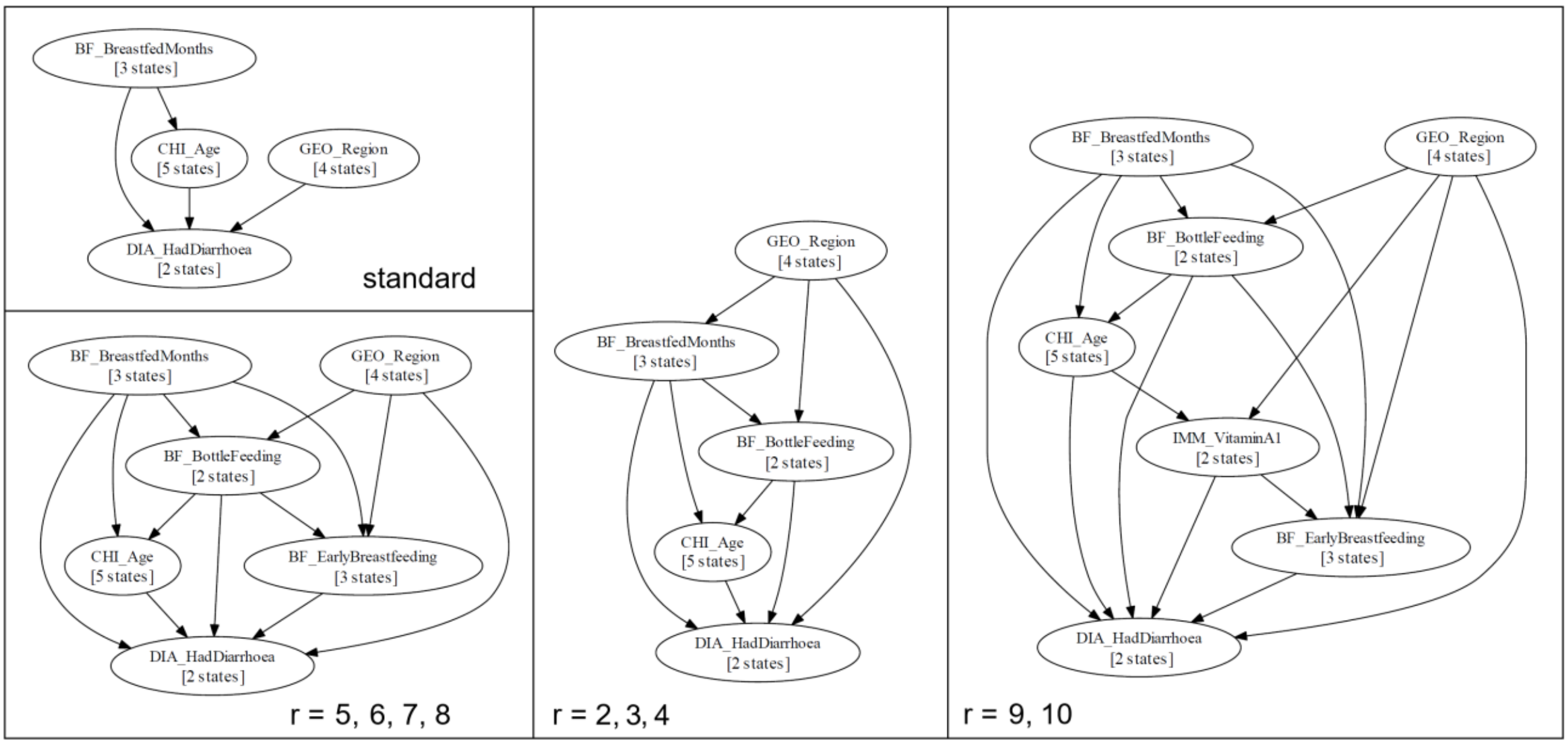

**Fig 12.** The Markov blankets of target variable *DIA_HadDiarrhoea* produced by GES given the childhood diarrhoea dataset, over different inputs $r_i$ while applying TAR-VAR (refer to Equation 2).

**Table 6.** The impact on target variable *Violence* from the forensic psychiatry dataset over different inputs $r_i$, its parameters, inbound links and nearby graph, where $p_u$ is the unweighted (i.e., ignores $r_i$) number of free parameters of *Violence*, $p_w$ is the weighted (i.e., assumes $r_i$) number of free parameters of *Violence*, $p_{wg}$ is the weighted number of free parameters of the whole graph $SHD_a$ is the SHD score between the graph generated at $r_i$ and the graph generated without a target (i.e., when $r_i = 1$), and $SHD_b$ is $SHD_a$ minus the additional parents of target variable *Violence* (i.e., it represents the number of graphical modifications occurred in the graph excluding the edges entering the target variable). Refer to Figs 7 to 9 for the corresponding Markov blankets of *Violence*.

| | HC | | | | | | |
|---|---|---|---|---|---|---|---|
| $r_i$ for Target *Violence* | $p_u$ | $p_w$ | $p_{wg}$ | $BIC_{TAR-VAR}$ | $SHD_a$ ($SHD_b$) | Parents of Target | Edges in graph |
| 1 | 4 | 4 | 433 | -51096.0 | 0 (0) | 1 | 76 |
| 2 | 16 | 8 | 437 | -51072.4 | 1 (0) | 2 | 77 |
| 3 | 36 | 12 | 441 | -51056.7 | 1 (0) | 2 | 77 |
| 4 | 36 | 9 | 438 | -51041.8 | 1 (0) | 2 | 77 |
| 5 | 36 | 7 | 436 | -51031.9 | 1 (0) | 2 | 77 |
| 6 | 144 | 24 | 453 | -51019.9 | 2 (0) | 3 | 78 |
| 7 | 144 | 20 | 449 | -51000.2 | 2 (0) | 3 | 78 |
| 8 | 144 | 18 | 447 | -50990.3 | 2 (0) | 3 | 78 |
| 9 | 144 | 16 | 445 | -50980.4 | 2 (0) | 3 | 78 |
| 10 | 288 | 28 | 457 | -50968.0 | 3 (0) | 4 | 79 |

| | TABU | | | | | | |
|---|---|---|---|---|---|---|---|
| $r_i$ for Target *Violence* | $p_u$ | $p_w$ | $p_{wg}$ | $BIC_{TAR-VAR}$ | $SHD_a$ ($SHD_b$) | Parents of Target | Edges in graph |
| 1 | 4 | 4 | 425 | -50972.2 | 0 (0) | 1 | 77 |
| 2 | 16 | 8 | 438 | -50919.1 | 5 (4) | 2 | 79 |
| 3 | 36 | 12 | 442 | -50903.4 | 5 (4) | 2 | 79 |
| 4 | 36 | 9 | 430 | -50918.2 | 1 (0) | 2 | 78 |
| 5 | 36 | 7 | 437 | -50878.6 | 5 (4) | 2 | 79 |
| 6 | 144 | 24 | 445 | -50896.3 | 2 (0) | 3 | 79 |
| 7 | 144 | 20 | 441 | -50876.5 | 2 (0) | 3 | 79 |
| 8 | 144 | 18 | 439 | -50866.7 | 2 (0) | 3 | 79 |
| 9 | 144 | 16 | 437 | -50856.8 | 2 (0) | 3 | 79 |
| 10 | 288 | 28 | 449 | -50844.4 | 3 (0) | 4 | 80 |

| | SaiyanH | | | | | | |
|---|---|---|---|---|---|---|---|
| $r_i$ for Target *Violence* | $p_u$ | $p_w$ | $p_{wg}$ | $BIC_{TAR-VAR}$ | $SHD_a$ ($SHD_b$) | Parents of Target | Edges in graph |
| 1 | 4 | 4 | 406 | -50994.1 | 0 (0) | 1 | 73 |
| 2 | 16 | 8 | 410 | -50970.7 | 1 (0) | 2 | 74 |
| 3 | 36 | 12 | 414 | -50955.0 | 1 (0) | 2 | 74 |
| 4 | 32 | 8 | 410 | -50950.4 | 2 (1) | 2 | 75 |
| 5 | 36 | 7 | 409 | -50930.2 | 1 (0) | 2 | 74 |
| 6 | 36 | 6 | 408 | -50925.3 | 1 (0) | 2 | 74 |
| 7 | 144 | 20 | 422 | -50907.6 | 2 (0) | 3 | 75 |
| 8 | 144 | 18 | 420 | -50897.7 | 2 (0) | 3 | 75 |
| 9 | 144 | 16 | 418 | -50887.8 | 2 (0) | 3 | 75 |
| 10 | 144 | 14 | 416 | -50877.9 | 2 (0) | 3 | 75 |

| | MAHC | | | | | | |
|---|---|---|---|---|---|---|---|
| $r_i$ for Target *Violence* | $p_u$ | $p_w$ | $p_{wg}$ | $BIC_{TAR-VAR}$ | $SHD_a$ ($SHD_b$) | Parents of Target | Edges in graph |
| 1 | 4 | 4 | 369 | -51361.2 | 0 (0) | 1 | 65 |
| 2 | 16 | 8 | 373 | -51337.9 | 2 (1) | 2 | 66 |
| 3 | 36 | 12 | 377 | -51322.1 | 1 (0) | 2 | 66 |
| 4 | 36 | 9 | 374 | -51307.3 | 1 (0) | 2 | 66 |
| 5 | 36 | 7 | 372 | -51297.4 | 1 (0) | 2 | 66 |
| 6 | 144 | 24 | 389 | -51285.4 | 2 (0) | 3 | 67 |
| 7 | 144 | 20 | 385 | -51265.6 | 2 (0) | 3 | 67 |
| 8 | 144 | 18 | 383 | -51255.7 | 3 (1) | 3 | 67 |
| 9 | 144 | 16 | 381 | -51245.8 | 2 (0) | 3 | 67 |
| 10 | 144 | 14 | 379 | -51235.9 | 2 (0) | 3 | 67 |

| | GES | | | | | | |
|---|---|---|---|---|---|---|---|
| $r_i$ for Target *Violence* | $p_u$ | $p_w$ | $p_{wg}$ | $BIC_{TAR-VAR}$ | $SHD_a$ ($SHD_b$) | Parents of Target | Edges in graph |
| 1 | 4 | 4 | 439 | -51122.007 | 0 (0) | 1 | 76 |
| 2 | 16 | 8 | 443 | -51098.625 | 1 (0) | 2 | 77 |
| 3 | 36 | 12 | 447 | -51082.862 | 1 (0) | 2 | 77 |
| 4 | 36 | 9 | 444 | -51068.018 | 1 (0) | 2 | 77 |
| 5 | 36 | 7 | 442 | -51058.121 | 1 (0) | 2 | 77 |
| 6 | 144 | 24 | 459 | -51046.156 | 2 (0) | 3 | 78 |
| 7 | 144 | 20 | 455 | -51026.363 | 2 (0) | 3 | 78 |
| 8 | 144 | 18 | 453 | -51016.467 | 2 (0) | 3 | 78 |
| 9 | 144 | 16 | 451 | -51006.570 | 2 (0) | 3 | 78 |
| 10 | 288 | 28 | 463 | -50994.204 | 3 (0) | 4 | 79 |

**Table 7.** The impact on target variable *DIA_HadDiarrhoea* from the childhood diarrhoea dataset over different inputs $r_i$, its parameters, inbound links and nearby graph, where $p_u$ is the unweighted (i.e., ignores $r_i$) number of free parameters of *DIA_HadDiarrhoea*, $p_w$ is the weighted (i.e., assumes $r_i$) number of free parameters of *DIA_HadDiarrhoea*, $p_{wg}$ is the weighted number of free parameters of the whole graph, $SHD_a$ is the SHD score between the graph generated at $r_i$ and the graph generated without a target (i.e., when $r_i = 1$), and $SHD_b$ is $SHD_a$ minus the additional parents of target variable *Violence* (i.e., it represents the number of graphical modifications occurred in the graph excluding the edges entering the target variable). Refer to Figs 10 to 12 for the corresponding Markov blankets of *DIA_HadDiarrhoea*.

| | HC | | | | | | | TABU | | | | | | | SaiyanH | | | | | | |
|---|---|---|---|---|---|---|---|---|---|---|---|---|---|---|---|---|---|---|---|---|---|
| $r_i$ for Target *DIA_HadDiarrhoea* | $p_u$ | $p_w$ | $p_{wg}$ | $BIC_{TAR-VAR}$ | $SHD_a$ ($SHD_b$) | Parents of Target | Edges in graph | $p_u$ | $p_w$ | $p_{wg}$ | $BIC_{TAR-VAR}$ | $SHD_a$ ($SHD_b$) | Parents of Target | Edges in graph | $p_u$ | $p_w$ | $p_{wg}$ | $BIC_{TAR-VAR}$ | $SHD_a$ ($SHD_b$) | Parents of Target | Edges in graph |
| 1 | 60 | 60 | 5194 | -6933326.6 | 0 (0) | 3 | 112 | 60 | 60 | 4942 | -6921437.5 | 0 (0) | 3 | 111 | 12 | 12 | 2525 | -6998323.6 | 0 (0) | 2 | 80 |
| 2 | 120 | 60 | 5194 | -6932995.9 | 1 (0) | 4 | 113 | 120 | 60 | 4942 | -6921106.8 | 1 (0) | 4 | 112 | 12 | 6 | 2519 | -6998269.7 | 0 (0) | 2 | 80 |
| 3 | 120 | 40 | 5174 | -6932816.0 | 1 (0) | 4 | 113 | 120 | 40 | 4922 | -6920926.9 | 1 (0) | 4 | 112 | 12 | 4 | 2517 | -6998251.7 | 0 (0) | 2 | 80 |
| 4 | 360 | 90 | 5224 | -6932716.0 | 2 (0) | 5 | 114 | 360 | 90 | 4972 | -6920826.9 | 2 (0) | 5 | 113 | 12 | 3 | 2516 | -6998242.7 | 0 (0) | 2 | 80 |
| 5 | 360 | 72 | 5206 | -6932554.1 | 2 (0) | 5 | 114 | 360 | 72 | 4954 | -6920665.0 | 2 (0) | 5 | 113 | 12 | 2 | 2515 | -6998233.7 | 0 (0) | 2 | 80 |
| 6 | 360 | 60 | 5194 | -6932446.2 | 2 (0) | 5 | 114 | 360 | 60 | 4942 | -6920557.1 | 2 (0) | 5 | 113 | 12 | 2 | 2515 | -6998233.7 | 0 (0) | 2 | 80 |
| 7 | 360 | 51 | 5185 | -6932365.2 | 2 (0) | 5 | 114 | 360 | 51 | 4933 | -6920476.1 | 2 (0) | 5 | 113 | 12 | 1 | 2514 | -6998224.7 | 0 (0) | 2 | 80 |
| 8 | 360 | 45 | 5179 | -6932311.3 | 2 (0) | 5 | 114 | 360 | 45 | 4927 | -6920422.2 | 2 (0) | 5 | 113 | 12 | 1 | 2514 | -6998224.7 | 0 (0) | 2 | 80 |
| 9 | 720 | 80 | 5214 | -6932221.4 | 3 (0) | 6 | 115 | 720 | 80 | 4962 | -6920332.3 | 3 (0) | 6 | 114 | 12 | 1 | 2514 | -6998224.7 | 0 (0) | 2 | 80 |
| 10 | 720 | 72 | 5206 | -6932149.4 | 3 (0) | 6 | 115 | 720 | 72 | 4954 | -6920260.3 | 3 (0) | 6 | 114 | 12 | 1 | 2514 | -6998224.7 | 0 (0) | 2 | 80 |

| | MAHC | | | | | | | GES | | | | | | |
|---|---|---|---|---|---|---|---|---|---|---|---|---|---|---|
| $r_i$ for Target *DIA_HadDiarrhoea* | $p_u$ | $p_w$ | $p_{wg}$ | $BIC_{TAR-VAR}$ | $SHD_a$ ($SHD_b$) | Parents of Target | Edges in graph | $p_u$ | $p_w$ | $p_{wg}$ | $BIC_{TAR-VAR}$ | $SHD_a$ ($SHD_b$) | Parents of Target | Edges in graph |
| 1 | 60 | 60 | 4503 | -6922301.7 | 0 (0) | 3 | 111 | 60 | 60 | 4961 | -6926663.9 | 0 (0) | 3 | 112 |
| 2 | 120 | 60 | 4503 | -6921971.0 | 2 (1) | 4 | 112 | 120 | 60 | 5021 | -6926872.8 | 1 (0) | 4 | 113 |
| 3 | 120 | 40 | 4483 | -6921791.1 | 2.5 (1.5) | 4 | 112 | 120 | 40 | 4941 | -6926153.316 | 1 (0) | 4 | 113 |
| 4 | 120 | 30 | 4473 | -6921701.2 | 2 (1) | 4 | 112 | 120 | 30 | 5261 | -6928481.402 | 2 (1) | 4 | 114 |
| 5 | 360 | 72 | 4515 | -6921529.2 | 3.5 (1.5) | 5 | 113 | 360 | 72 | 4973 | -6925891.406 | 2 (0) | 5 | 114 |
| 6 | 360 | 60 | 4503 | -6921421.3 | 2.5 (0.5) | 5 | 113 | 360 | 60 | 4961 | -6925783.490 | 2 (0) | 5 | 114 |
| 7 | 360 | 51 | 4494 | -6921340.4 | 3 (1) | 5 | 113 | 360 | 51 | 4952 | -6925702.552 | 2 (0) | 5 | 114 |
| 8 | 360 | 45 | 4488 | -6921286.4 | 3 (1) | 5 | 113 | 360 | 45 | 4946 | -6925648.594 | 2 (0) | 5 | 114 |
| 9 | 720 | 80 | 4523 | -6921196.5 | 4 (1) | 6 | 114 | 720 | 80 | 4981 | -6925558.671 | 3 (0) | 6 | 115 |
| 10 | 720 | 72 | 4515 | -6921124.6 | 3.5 (0.5) | 6 | 114 | 720 | 72 | 4973 | -6925486.727 | 3 (0) | 6 | 115 |

### 6.3. *Evaluation of knowledge approaches* REL-BDN *and* STR-BDN

Lastly, the effect of approaches REL-BDN and STR-BDN is illustrated graphically in Fig 13, based on the Asia network with a sample size of $10^4$, and with application to the five algorithms HC, TABU, SaiyanH, MAHC and GES. BDNs are often developed manually and may include decision nodes not available in the data. However, it is possible for both decision and utility variables to be present in the data, and approaches REL-BDN and STR-BDN become useful under this scenario. With reference to Fig 13, the graphs:

a) in the first row represent the standard graphical output generated by each algorithm in the absence of constraints;
b) in the second row represent the graphical outputs after applying REL-BDN to each of the algorithms, where in this hypothetical example node *asia* serves as the decision node and node *smoke* as the utility node;
c) in the third row represent the graphical outputs after applying STR-BDN to each of the algorithms, for the same decision and utility nodes as in REL-BDN.

Note that nodes *asia* and *smoke* are constrained as decision and utility nodes respectively, purely for illustration purposes. As can be seen in Fig 13, the modifications applied by REL-BDN (second row) are restricted to visual representation and do not affect the graphical structure of the graph. Approach REL-BDN simply involves automatically reformatting the graphical output to incorporate knowledge of decisions and utilities present in the data, and converting conditional arcs entering decisions into informational arcs. On the other hand, approach STR-BDN (third row) constrains structure learning by requiring that decision nodes have at least one child, and utility nodes have at least one parent. These constraints are based on the assumption that decision nodes must have at least an effect node in the dataset, and utility nodes must have at least one cause in the data.

For example, Fig 13 shows that HC with REL-BDN produces a graph in which decision node *asia* has no child, and this can be a problem in practice when we automatically convert learnt graphs into parameterised[8] BDNs, since the specified decision node will have no effect in the network. On the other hand, the same HC algorithm ensures that node *asia* does have a child node when paired with STR-BDN. Note that while in the case of HC and SaiyanH this was achieved by introducing just one additional arc, in the case of TABU and MAHC this constraint was satisfied after adding two arcs. This is because STR-BDN forces the algorithm to keep adding arcs that minimally decrease the BIC score until both constraints are satisfied. Because the arc that minimises BIC will not necessarily be the arc that is needed for the constraint to be satisfied, multiple arcs may have to be introduced before the constraint is met, as in the example with TABU and MAHC.

[8] Our implementation has been extended to include parameterisation of the learnt BDN graphs into BDN models that have the appropriate file extensions to be loaded in the third-party BN software engines AgenaRisk and GeNIe. However, this is out of the scope of this paper. Details can be found in the Bayesys user manual (Constantinou, 2019).

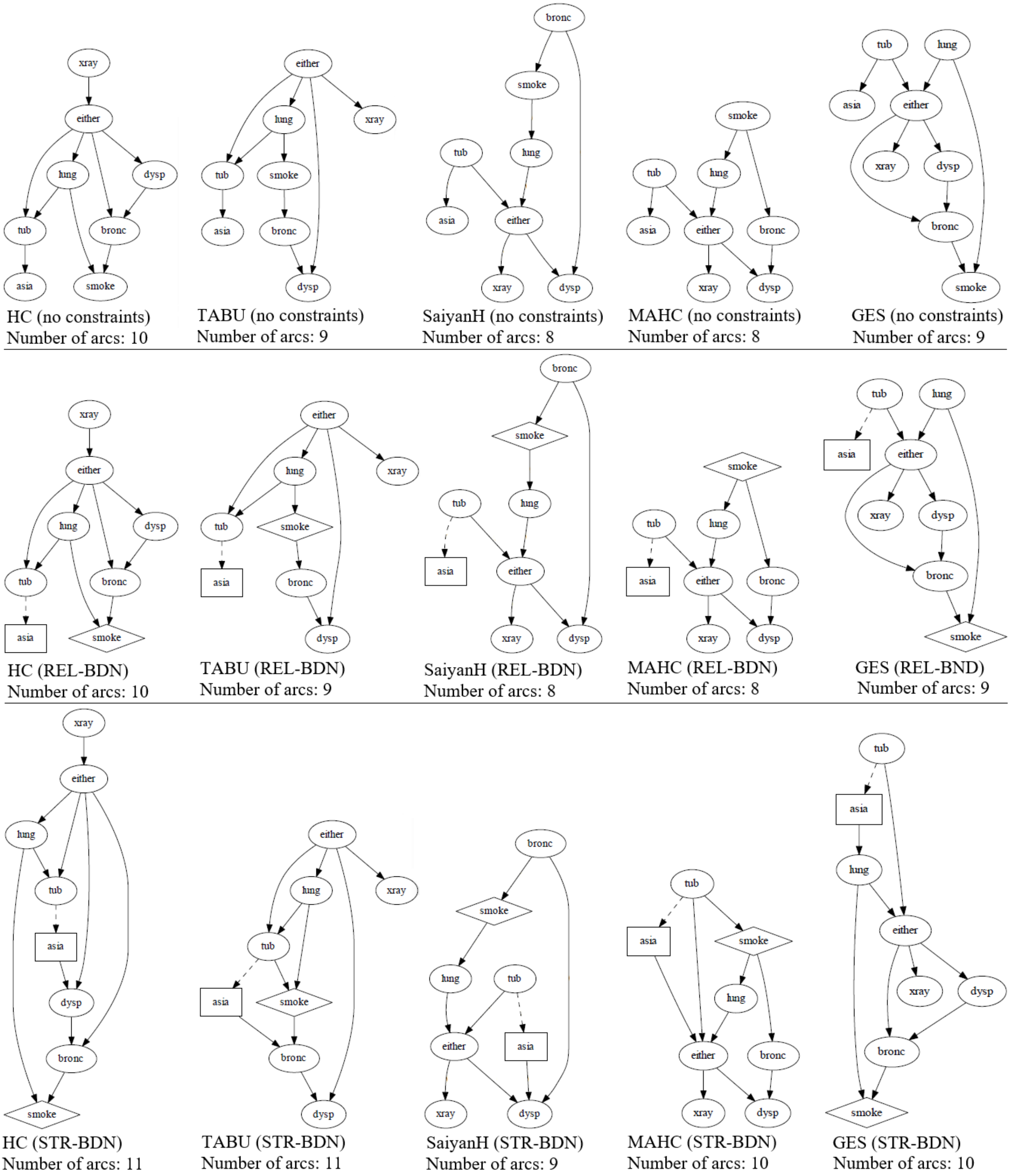

**Fig 13.** The learnt Asia graphs produced by the five algorithms a) in the absence of constraints shown on the left, b) given approach REL-BDN with nodes *asia* and *smoke* set as decision and utility nodes respectively shown in the middle, and c) given approach STR-BDN for the same decision and utility nodes shown on the right.

## 7. Concluding remarks

This paper introduced some novel and some existing knowledge approaches. In total, we have implemented and evaluated 10 knowledge approaches that impose soft or hard constraints on BNSL algorithms. These knowledge approaches enable users to perform BNSL by considering both data and knowledge that could be obtained from heterogeneous sources with different conceptual, contextual and typographical representation. The insights obtained through evaluation can be used to inform decision making about which type of knowledge might be most beneficial and under what data settings, and to answer the research questions discussed in Section 1.

The implementations are made available in the open-source Bayesys structure learning system [14]. The knowledge approaches have been tested by implementing them within the process of five structure learning algorithms available in Bayesys. We have evaluated the effectiveness and efficiency of these approaches by testing their impact on structure learning in terms of graphical accuracy, model fitting, complexity, and runtime, over various synthetic and real datasets with varied levels of complexity and sample size. To the best of our knowledge, this paper provides the first broad evaluation of different knowledge approaches for BNSL.

The results show that, amongst the knowledge approaches investigated, the most useful in terms of improving the graphical accuracy of the learnt graph is approach DIR-EDG (*Directed edge*) that captures information about the existence and direction of edges. DIR-EDG not only represents one of the most widely used knowledge approaches in the literature, but also represents one of the simplest approaches. As illustrated in Fig 14 which provides a concluding summary, this is also true when we plot the gain in graphical accuracy relative to the number of edges or variables specified as constraints (i.e., the pieces of knowledge) imposed by the different knowledge approaches; although in this case the value of knowledge through the INI-GRA (*Initial graph*) approach is only marginally behind DIR-EDG (*Directed Edge*).

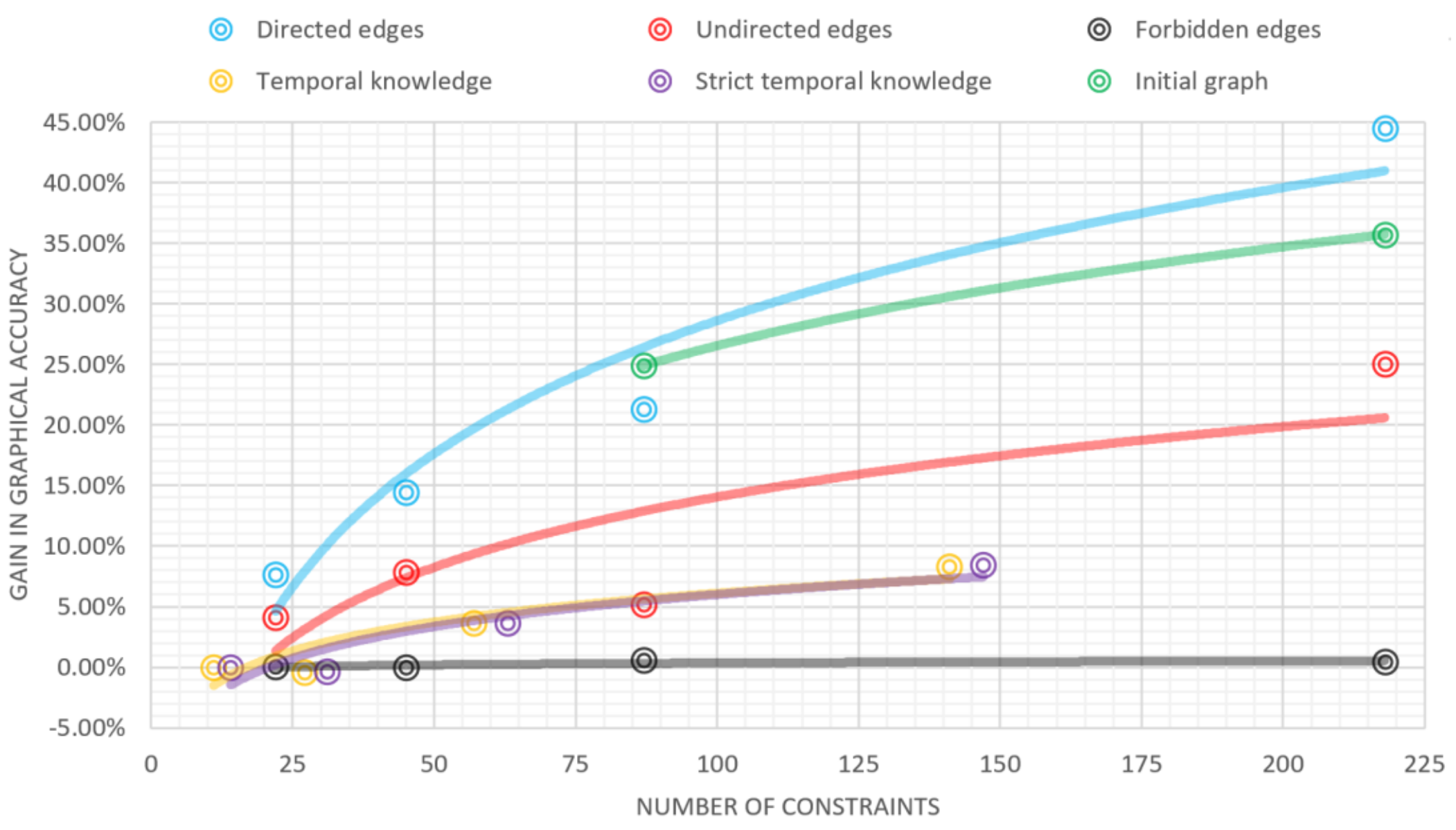

**Fig 14.** Concluding summary of the gain in graphical accuracy relative to the number of edges or variables specified as constraints imposed by each knowledge approach, as an average across all case studies and algorithms. In this plot, the graphical accuracy represents the average logarithmic fit across the F1, BSF and SHD metrics.

Furthermore, the results show that while most approaches become less effective in the presence of big data, presumably because learning accuracy increases with sample size thereby rendering knowledge less important, some approaches perform better still in the presence of big data. While this is counterintuitive, it is explained by the higher number of edges the algorithms generate when the input data are 'big'. In these cases, approaches that forbid edges, for example, can help algorithms avoid discovering false positives edges; a phenomenon that becomes more prevalent with increasing sample size [34].

Amongst the main conclusions is the observation that reduced search space due to knowledge does not imply reduced computational complexity. In fact, our results suggest that in many cases the incorporation of knowledge increases runtime. While the increase in runtime can partly be explained by the additional computation needed to apply and search for constraints, it also suggests that the number of graphs explored increases, or that the actual graphs explored in the presence of constraints are more complex (i.e., contain a higher number of free parameters) which require a higher computational effort to be scored. This can happen when the constraints set up a tension between what the data indicate and what the constraints are trying to enforce, and which becomes more likely when the sample size of the input data is limited, irrespective of the accuracy of the constraints. As the BIC score comparisons show (refer to Fig 5), incorporating knowledge tends to decrease score fitting while at the same time increases graphical accuracy. This phenomenon is particularly evidenced in the presence of limited data, where objective functions such as the BIC tend to be less effective.

This study comes with some limitations, all of which can also be viewed as directions for future research. Regarding the density and size of the networks, the analysis is based on networks containing up to hundreds of nodes, hundreds of edges, with an average degree up to 3.58, a maximum in-degree up to 6, and up to 71,890 free parameters. While we consider the selected networks to be well within the complexity range of most knowledge-based BNs found in the literature, our analysis did not consider denser networks containing thousands of variables, which are typically found in bioinformatics. These larger networks are out of the scope of this paper for two reasons. Firstly, networks of this size require specialised algorithms that maximise learning efficiency, often at the expense of accuracy, which are not considered in this paper. Secondly, and perhaps more importantly, it is well documented that knowledge elicitation can be time consuming even when applied to networks containing tens of variables. On this basis, we assume that it would be impractical to expect users to elicit knowledge for networks containing thousands of variables. Moreover, the impact of knowledge is investigated by applying the different knowledge approaches to the five algorithms that are available in Bayesys; three score-based algorithms, a model-averaging score-based algorithm, and a hybrid learning algorithm. While most of the results appear to be rather consistent across algorithms, we cannot claim that these results extend to all other algorithms. Lastly, all algorithms were employed using the BIC score as the objective function to ensure consistency in measuring the impact on objective scores. While the algorithms are found to behave similarly irrespective of the selection of the objective function [35], some algorithms may be more sensitive than others to the objective score in the presence of limited data [44].

It is important to highlight that knowledge elicitation comes with disadvantages, such as requiring access to expertise which tends to increase effort, cost, and project duration. Moreover, because knowledge can be subjective and inconsistent between experts, some problems may require access to multiple domain experts to reduce the uncertainty of knowledge elicited. Because of these reasons, we have avoided implementing approaches that require experts to provide subjective probabilities as prior beliefs, such as those covered in [1, 6, 7, 28, 37], and this can be viewed as a limitation of this study.

Lastly, most of the results presented in this study assume constraints sampled from ground truth graphs. While this assumption is in line with some other past studies, it influences the results in two conflicting ways. Firstly, because knowledge is sampled from the error-free ground truth graphs, their positive effect tends to be overestimated. Secondly, and in contrast to the first point, because structure learning is based on synthetic data which is known to overestimate real-world performance, it tends to underestimate the benefits of knowledge since higher learning accuracy renders knowledge less useful. While it is not possible to know the relationship between these two conflicting effects, there is clearly a cancelling effect. Still, the results presented in this paper provide us with valuable insights in terms of how the impact of one knowledge approach relates to another, irrespective of the underlying knowledge and data assumptions discussed above.

## Acknowledgements

This research was supported by the EPSRC Fellowship project EP/S001646/1 on *Bayesian Artificial Intelligence for Decision Making under Uncertainty*.

## Appendix A: Supplementary description for SaiyanH

The material that follows is based on the description of SaiyanH presented in [15]. The Mean/Max/MeanMax Marginal Discrepancy ($MMD$) score represents the discrepancy in marginal probabilities between prior and posterior distributions, and ranges from 0 to 1 where a higher score indicates a stronger dependency. For edge $A \leftrightarrow B$, the score $MMD(A \leftrightarrow B)$ is the average of scores $MMD_{MN}(A \leftrightarrow B)$ and $MMD_{MX}(A \leftrightarrow B)$, where $MN$ and $MX$ are mean and max marginal discrepancies. Specifically,

$$MMD(A \leftrightarrow B) = \sum_{\leftrightarrow} \sum_{m} MMD_m(A \leftrightarrow B)w$$

where $\leftrightarrow$ represents the iterations over $\leftarrow$ and $\rightarrow$, $m$ represents the iterations over $MN$ and $MX$, and $w$ is the normalising constant 0.25 applied to the scores accumulated over the following four iterations:

$$MMD_{MN}(A \rightarrow B) = \left( \sum_{j}^{s_A} \left[ \left( \sum_{i}^{s_B} \left| P(B_i) - P(B_i|A_j) \right| \right) \Big/ s_B \right] \right) \Big/ S_A$$

$$MMD_{MN}(A \leftarrow B) = \left( \sum_{i}^{s_B} \left[ \left( \sum_{j}^{s_A} \left| P(A_j) - P(A_j|B_i) \right| \right) \Big/ S_A \right] \right) \Big/ s_B$$

$$MMD_{MX}(A \rightarrow B) = \left( \sum_{j}^{s_A} \max_{i} \left| P(B_i) - P(B_i|A_j) \right| \right) \Big/ S_A$$

$$MMD_{MX}(A \leftarrow B) = \left( \sum_{i}^{s_B} \max_{j} \left| P(A_j) - P(A_j|B_i) \right| \right) \Big/ s_B$$

for each state $j$ in $A$ and state $i$ in $B$, and over the $S_A$ states in $A$ and $S_B$ states in $B$.

The Extended Maximum Spanning Graph (EMSG) is determined by the $MMD$ scores described above, and can be viewed a more complex version of the maximum spanning tree that preserves multiple connecting paths from one node to another (unlike the maximum spanning tree which preserves the single and most likely connecting path between nodes). Starting from a complete graph, the EMSG is produced by removing edges between two nodes $A$ and $B$ if and only if $A$ and $B$ share neighbour $C$ where

$$MMD(A \leftrightarrow C) > MMD(A \leftrightarrow B) < MMD(B \leftrightarrow C)$$

The order in which the edges are assessed for removal is from lowest to highest $MMD$ score.

## Appendix B: Supplementary results

**Table B.1.** The relative impact knowledge approaches have on structure learning performance, per algorithm, in terms of F1, BSF and SHD scores (overall), where DIR-EDG is directed edge constraints, UND-EDG is undirected edge constraints, FOR-EDG is forbidden edges, REL-TEM is relaxed temporal order, STR-TEM is strict temporal order, INI-GRA is input graph, and VAR-REL is variables are relevant. Positive percentages represent improvements.

| Approach | DIR-EDG | | | | UND-EDG | | | | FOR-EDG | | | | REL-TEM | | | | STR-TEM | | | | INI-GRA | | VAR-REL |
|---|---|---|---|---|---|---|---|---|---|---|---|---|---|---|---|---|---|---|---|---|---|---|---|
| Rate | 5% | 10% | 20% | 50% | 5% | 10% | 20% | 50% | 5% | 10% | 20% | 50% | 5% | 10% | 20% | 50% | 5% | 10% | 20% | 50% | 20% | 50% | |
| Measure | F1 | | | | | | | | | | | | | | | | | | | | | | |
| Impact on HC | 14% | 29% | 42% | 83% | 8% | 19% | 25% | 51% | 0% | 0% | 1% | 0% | 0% | 2% | 6% | 17% | 0% | 2% | 6% | 18% | 60% | 72% | -2% |
| Impact on TABU | 17% | 25% | 39% | 75% | 9% | 14% | 24% | 48% | 0% | 0% | 1% | -1% | 0% | -3% | 11% | 18% | 0% | -3% | 11% | 17% | 45% | 60% | -1% |
| Impact on SaiyanH | 4% | 8% | 15% | 41% | 2% | 4% | 10% | 30% | 0% | 0% | -1% | 1% | 0% | 0% | 2% | 5% | 0% | 0% | 2% | 6% | 22% | 53% | n/a |
| Impact on MAHC | 4% | 10% | 17% | 40% | 3% | 5% | 9% | 24% | 0% | 0% | 1% | 1% | 0% | -1% | 0% | 3% | 0% | -1% | 0% | 3% | 7% | 15% | -2% |
| Impact on GES | 9% | 13% | 22% | 48% | 0% | 3% | 4% | 7% | 0% | 0% | 0% | 0% | 0% | 0% | 0% | 11% | 0% | 0% | 0% | 11% | 14% | 29% | -2% |
| Overall impact | 10% | 17% | 27% | 57% | 4% | 9% | 15% | 32% | 0% | 0% | 0% | 0% | 0% | -1% | 4% | 11% | 0% | 0% | 4% | 11% | 30% | 46% | -2% |
| Measure | BSF | | | | | | | | | | | | | | | | | | | | | | |
| Impact on HC | 28% | 63% | 77% | 142% | 20% | 43% | 55% | 95% | 0% | 0% | 5% | 5% | 0% | 2% | 20% | 33% | 0% | 2% | 20% | 34% | 103% | 122% | -4% |
| Impact on TABU | 27% | 38% | 52% | 98% | 17% | 20% | 37% | 68% | 0% | 0% | 3% | 0% | 0% | -3% | 15% | 24% | 0% | -3% | 15% | 23% | 59% | 78% | -3% |
| Impact on SaiyanH | 3% | 10% | 18% | 50% | 2% | 6% | 13% | 38% | 0% | 0% | -1% | 1% | 0% | -1% | 2% | 5% | 0% | 0% | 2% | 6% | 26% | 63% | n/a |
| Impact on MAHC | 4% | 11% | 19% | 48% | 3% | 6% | 11% | 31% | 0% | 0% | 1% | 1% | 0% | -1% | 0% | 2% | 0% | -1% | 0% | 2% | 8% | 18% | 1% |
| Impact on GES | 6% | 12% | 20% | 45% | -1% | 2% | 5% | 9% | 0% | 0% | 0% | 0% | 0% | 0% | 0% | 9% | 0% | 0% | 0% | 9% | 12% | 22% | 0% |
| Overall impact | 14% | 27% | 37% | 77% | 8% | 15% | 24% | 48% | 0% | 0% | 1% | 1% | 0% | -1% | 7% | 15% | 0% | -1% | 8% | 15% | 42% | 61% | -1% |
| Measure | SHD | | | | | | | | | | | | | | | | | | | | | | |
| Impact on HC | -8% | -18% | -28% | -64% | -3% | -10% | -13% | -33% | 0% | 0% | -1% | -1% | 0% | -3% | -5% | -12% | 0% | -3% | -5% | -13% | -50% | -64% | 15% |
| Impact on TABU | -10% | -13% | -24% | -61% | -5% | -5% | -10% | -31% | 0% | 0% | -2% | 2% | 1% | 9% | -10% | -12% | 1% | 8% | -10% | -11% | -34% | -45% | 11% |
| Impact on SaiyanH | -3% | -10% | -17% | -48% | -2% | -5% | -11% | -37% | -1% | -1% | 1% | -2% | 1% | 1% | -3% | -7% | 1% | 0% | -3% | -8% | -24% | -53% | n/a |
| Impact on MAHC | -4% | -10% | -19% | -47% | -2% | -5% | -10% | -25% | 0% | 0% | -1% | -1% | 0% | 2% | 1% | -3% | 0% | 2% | 1% | -4% | -10% | -21% | 26% |
| Impact on GES | -11% | -21% | -28% | -57% | 3% | -5% | 11% | 49% | 0% | 0% | 0% | 0% | 0% | 0% | 0% | -13% | 0% | 0% | 0% | -13% | -22% | -38% | 15% |
| Overall impact | -7% | -14% | -23% | -56% | -2% | -6% | -7% | -15% | 0% | 0% | 0% | 0% | 0% | 2% | -3% | -10% | 0% | 2% | -3% | -10% | -28% | -44% | 17% |
| Measure | BIC | | | | | | | | | | | | | | | | | | | | | | |
| Impact on HC | 0% | 0.1% | 0.1% | 0% | -0.2% | 0.1% | 0% | 0% | 0% | 0% | 0% | 0% | 0% | 0% | -0.6% | 0.1% | 0% | 0% | -0.7% | 0% | 0.2% | 0.2% | 0.0% |
| Impact on TABU | 0% | -0.1% | -0.4% | -1.3% | 0.1% | -0.2% | -0.4% | -0.9% | 0% | 0% | 0% | 0% | 0% | 0% | -0.1% | 0.1% | 0% | 0% | 0% | 0.1% | 14.5% | 0.0% | 0.0% |
| Impact on SaiyanH | -0.2% | -0.2% | -0.1% | -0.7% | 0% | 0% | 0.2% | 0% | 0% | 0% | 0% | 0.1% | 0% | 0% | 0.1% | -0.3% | 0% | 0% | 0.1% | -0.6% | 0.7% | 0.7% | n/a |
| Impact on MAHC | -0.1% | -0.4% | -0.5% | -0.9% | -0.2% | -1.0% | -101.4% | -10.8% | 0% | 0% | 0% | 0% | 0.1% | -0.1% | -0.3% | -1.0% | 0.1% | -0.1% | -0.3% | -1.5% | -0.1% | 0.3% | 0.3% |
| Impact on GES | 0.1% | 0.1% | 0% | -0.1% | 0% | -3.4% | -14.5% | -14.3% | 0% | 0% | 0% | 0% | 0% | 0% | -0.5% | 0% | 0% | 0% | -0.5% | 0% | 0.1% | 0.1% | 0.0% |
| Overall impact | 0% | -0.1% | -0.2% | -0.6% | -0.1% | -0.9% | -23.2% | -5.2% | 0% | 0% | 0% | 0% | 0% | 0% | -0.3% | -0.2% | 0% | 0% | -0.3% | -0.4% | 3.1% | 0.3% | 0.1% |

| Measure | Free parameters | | | | | | | | | | | | | | | | | | | | | | |
|---|---|---|---|---|---|---|---|---|---|---|---|---|---|---|---|---|---|---|---|---|---|---|---|
| Impact on HC | 0% | -12% | 0% | 0% | 4% | -12% | 0% | 7% | 0% | 0% | 4% | 3% | 0% | -1% | 16% | -5% | 0% | -1% | 15% | -5% | -22% | -28% | 2% |
| Impact on TABU | -9% | -1% | 10% | 34% | -5% | 6% | 17% | 35% | 0% | 0% | -4% | -3% | 5% | 4% | -4% | -6% | 5% | 4% | -1% | -8% | -11% | -4% | 2% |
| Impact on SaiyanH | 5% | 15% | 22% | 63% | 3% | 7% | 14% | 45% | 0% | 0% | -1% | 1% | 4% | 2% | 2% | 3% | 4% | 3% | 1% | 10% | 21% | 70% | n/a |
| Impact on MAHC | 3% | 7% | 15% | 56% | 3% | 28% | 3242% | 348% | 0% | 0% | -2% | -2% | 0% | 1% | -3% | 0% | 0% | 1% | -3% | 0% | -1% | 8% | 11% |
| Impact on GES | -9% | -5% | 1% | 19% | 0% | 223% | 1013% | 1321% | 0% | 0% | 0% | 2% | 7% | 3% | 4% | 1% | 7% | 3% | 4% | 1% | -6% | -11% | 3% |
| Overall impact | -2% | 1% | 9% | 34% | 1% | 50% | 857% | 351% | 0% | 0% | -1% | 0% | 3% | 2% | 3% | -1% | 3% | 2% | 3% | 0% | -4% | 7% | 5% |
| Measure | Arcs learnt | | | | | | | | | | | | | | | | | | | | | | |
| Impact on HC | -2% | -4% | -3% | -2% | 27% | -2% | 2% | 7% | 0% | 0% | -1% | -1% | 0% | -1% | 0% | -4% | 0% | -1% | -2% | -4% | -10% | -14% | 13% |
| Impact on TABU | -2% | 0% | 0% | 1% | -2% | 3% | 4% | 8% | 0% | 0% | -1% | -1% | 0% | 1% | -3% | -4% | 0% | 1% | -3% | -4% | -3% | 0% | 9% |
| Impact on SaiyanH | 0% | 0% | 2% | 7% | 0% | 0% | 2% | 6% | 0% | 0% | -2% | -1% | 0% | -1% | -1% | -1% | 0% | -1% | -1% | 1% | 1% | 5% | n/a |
| Impact on MAHC | 1% | 2% | 4% | 15% | 2% | 3% | 5% | 16% | 0% | 0% | -1% | -1% | 0% | 0% | -2% | -5% | 0% | 0% | -2% | -6% | 1% | 4% | 25% |
| Impact on GES | -4% | -3% | -2% | 1% | -1% | 0% | 6% | 25% | 0% | 0% | 0% | 0% | 0% | 0% | 1% | -3% | 0% | 0% | 1% | -3% | -6% | -13% | 13% |
| Overall impact | -1% | -1% | 0% | 4% | 5% | 1% | 4% | 13% | 0% | 0% | -1% | -1% | 0% | 0% | -1% | -3% | 0% | 0% | -1% | -3% | -3% | -4% | 15% |
| Measure | Runtime | | | | | | | | | | | | | | | | | | | | | | |
| Impact on HC | -5% | -9% | -12% | -31% | -4% | -8% | -2% | -8% | -4% | -1% | -1% | -3% | 0% | 13% | 48% | 214% | 3% | 10% | 44% | 190% | -26% | -46% | 0% |
| Impact on TABU | 35% | 24% | 25% | 22% | 25% | 24% | 15% | 18% | -1% | -6% | -3% | 23% | 15% | 40% | 126% | 195% | 14% | 30% | 141% | 206% | 21% | -21% | 0% |
| Impact on SaiyanH | 0% | -3% | -3% | -3% | 0% | 0% | 2% | -1% | 0% | 0% | 1% | 1% | -1% | 1% | 1% | 2% | -1% | 0% | 0% | 2% | -3% | -5% | n/a |
| Impact on MAHC | 0% | 0% | 0% | 6% | 1% | 3% | 1% | 1% | -3% | -3% | -4% | -4% | -1% | -1% | -1% | -1% | 0% | 4% | 2% | 2% | 2% | 2% | 2% |
| Impact on GES | -28% | -29% | -29% | -41% | -19% | -17% | -24% | -23% | -15% | -7% | -19% | -15% | -26% | -15% | -2% | 39% | -13% | -5% | 3% | 42% | -13% | -32% | -12% |
| Overall impact | 1% | -4% | -4% | -10% | 0% | 1% | -2% | -3% | -4% | -3% | -5% | 0% | -2% | 8% | 34% | 90% | 1% | 8% | 38% | 88% | -4% | -21% | -3% |

## Appendix C: Impact of knowledge on graphical scores

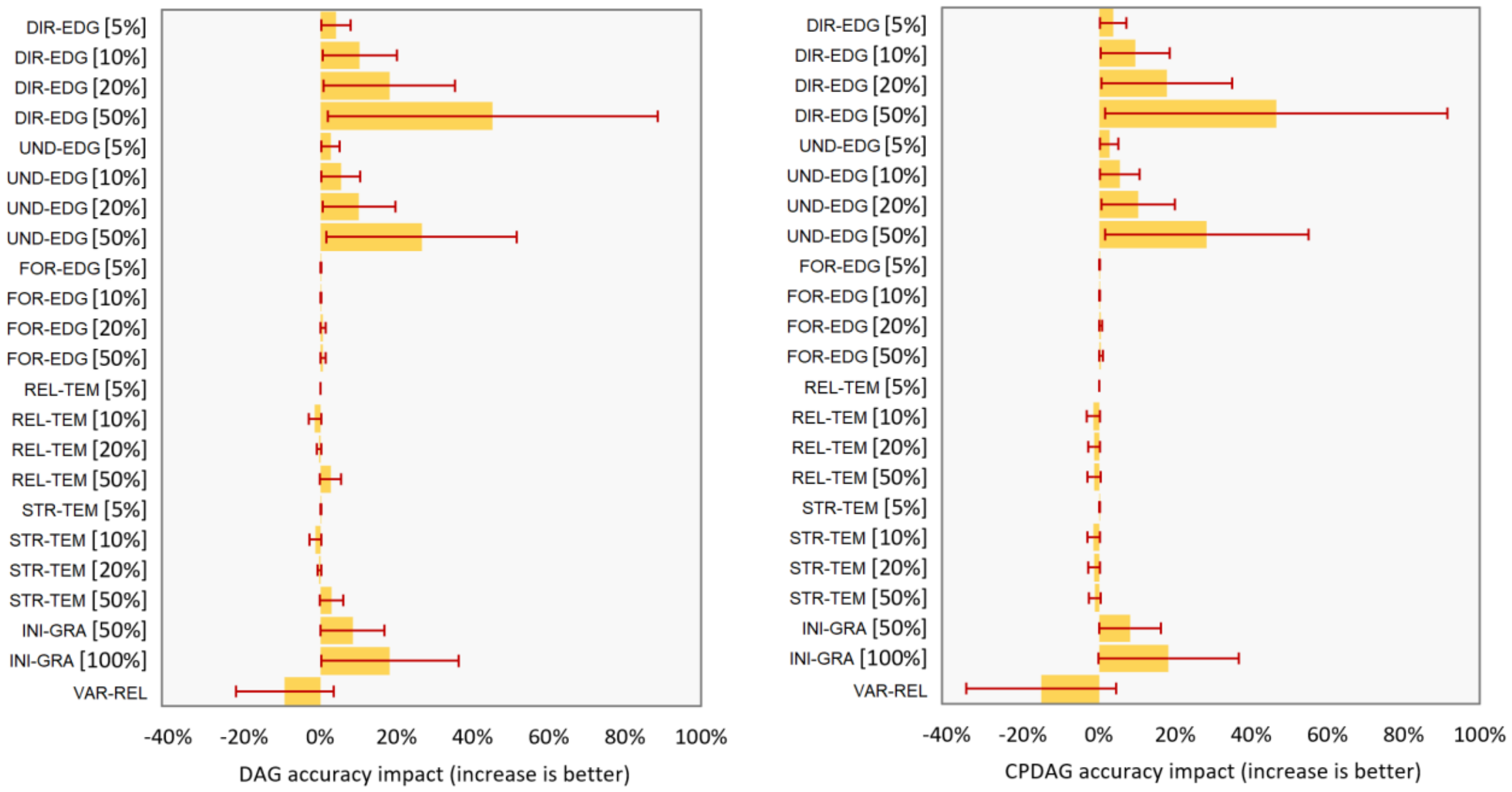

**Figure C.1.** The relative impact each knowledge approach has on structure learning performance, and over different rates of constraint, in terms of average graphical accuracy representing the average across F1, BSF and (invert) SHD scores. The graph on the left represents the DAG score results, and the graph on the right the CPDAG score results. Error lines represent standard deviation.

## Appendix D: Glossary

**Table D.1.** Glossary of abbreviations, symbols, and knowledge approaches.

| Abbreviation/Symbol/ Knowledge approach | Description |
|---|---|
| BDN | Bayesian Decision Network. |
| BIC | Bayesian Information Criterion. |
| BN | Bayesian Network. |
| BNSL | Bayesian Network Structure Learning. |
| BSF | Balanced Scoring Function. |
| CBN | Causal Bayesian Network. |
| CPDAG | Completed Partial Directed Acyclic Graph. |
| CPS | Candidate Parent Set. |
| DAG | Directed Acyclic Graph. |
| EMSG | Extended Maximum Spanning Graph. |
| GES | Greedy Equivalence Search. |
| HC | Hill Climbing. |
| ID | Influence Diagram. |
| LL | Log-Likelihood. |
| MAHC | Model-Averaging Hill-Climbing. |
| SHD | Structural Hamming Distance. |
| $A \rightarrow B$ | Causal relationship or direct dependency from variable $A$ to variable $B$. |
| $A - B$ | Causal relationship or direct dependency between variables $A$ and $B$ without knowledge of the direction of the relationship. |
| $A \perp B$ | No causal relationship or direct dependency between variables $A$ and $B$. |
| $N$ | Sample size. |
| $p$ | Number of free parameters. |
| $G$ | A given graph. |
| $V$ | The set of variables in graph $G$. |
| $\lvert V\rvert$ | The size of set $V$. |
| $s_i$ | The number of states of $v_i$. |
| $\pi_{v_i}$ | The parent set of $v_i$. |
| $\lvert\pi_{v_i}\rvert$ | The size of set $\pi_{v_i}$. |
| $q_j$ | The number of states of $v_j$ in parent set $\pi_{v_i}$. |
| $r_i$ | The penalty on free parameters for targeted variables. |
| $D$ | A given dataset. |
| $G_{max}$ | The graph that maximises a given score function. |
| $S_G$ | The score of $G$. |
| $S_{max}$ | The highest score across all graphs visited. |
| $G_n$ | The valid neighbouring graphs of $G$. |
| $S_{G_n}$ | The objective score of a given neighbouring graph. |
| $\vDash$ | Satisfies. |
| $K$ | A set of knowledge constraints. |
| $M$ | Maximum node in-degree. |
| $M_P$ | Maximum node in-degree during pruning. |
| $M_S$ | Maximum node in-degree during structure learning. |
| $MMD$ | Mean/Max/MeanMax Marginal Discrepancy. |
| $MN$ | Mean marginal discrepancy. |
| $MX$ | Mac marginal discrepancy. |
| $\overline{S(G,G_n)}$ | The average objective score of graph G and all its neighbouring graphs $G_n$. |
| $\boldsymbol{S}$ | A set of objective scores. |
| $F$ | A set of edges pruned off. |
| $a$ | Number of edges present in the true graph. |
| $i$ | Number of edges absent in the true graph. |
| $TP$ | True Positives. |
| $TN$ | True Negatives. |
| $FP$ | False Positives. |
| $FN$ | False Negatives. |
| DIR-EDG | Directed edge. |
| FOR-EDG | Forbidden edge. |
| INI-GRA | Initial graph. |
| REL-BDN | Relaxed Bayesian Decision Network. |
| REL-TEM | Relaxed incomplete temporal order. |
| STR-BDN | Strict Bayesian Decision Network. |
| STR-TEM | Strict incomplete temporal order. |
| TAR-VAR | Target variable. |
| UND-EDG | Undirected edge. |
| VAR-REL | Variables are relevant. |